\documentclass[runningheads]{llncs}
\usepackage{graphicx}
\usepackage{todonotes}
\usepackage{subfiles}
\usepackage{caption}
\usepackage{subcaption}
\usepackage{hyperref}
\usepackage{booktabs}
\hypersetup{
    colorlinks=true,
    linkcolor=blue,
    filecolor=magenta,      
    urlcolor=cyan,
}
\usepackage{amsmath}
\usepackage{adjustbox}
\usepackage{pdfpages}
\usepackage{booktabs}
\begin{document}
%
% \title{Contribution Title\thanks{Supported by organization x.}}
%\title{War of the Annotators}%Disease-specific Atlas Segmentation}
\title{Can non-specialists provide high quality gold standard labels in challenging modalities?}
\titlerunning{Are novice annotators good enough?}
\author{Samuel Budd\inst{1}\orcidID{0000-0002-9062-0013},
Thomas Day\inst{2,3},
John Simpson\inst{2,3},
Karen Lloyd\inst{2},
Jacqueline Matthew\inst{2,3},
Emily Skelton\inst{2,3,4},
Reza Razavi\inst{2,3},
Bernhard Kainz\inst{1,5}\orcidID{0000-0002-7813-5023}
}
% index{Budd, Samuel}
% index{Day, Thomas}
% index{Simpson, John}
% index{Lloyd, Karen}
% index{Matthew, Jacqueline}
% index{Skelton, Emily}
% index{Razavi, Reza}
% index{Kainz, Bernhard}
\authorrunning{S. Budd et al}
\institute{Imperial College London, Dept. Computing, BioMedIA, London, UK \and
King’s College London, London, UK \and
Guy’s and St Thomas’ NHS Foundation Trust, London, UK \and
School of Health Sciences, City, University of London, London, UK \and
Friedrich--Alexander University Erlangen--N\"urnberg, DE\\
\email{samuel.budd13@imperial.ac.uk}}
%King's College London, ISBE, London, UK \\ \email{samuel.budd13@imperial.ac.uk} }
\maketitle              % typeset the header of the contribution
\begin{abstract}
Probably yes. --- Supervised Deep Learning dominates performance scores for many computer vision tasks and defines the state-of-the-art. However, medical image analysis lags behind natural image applications. One of the many reasons is the lack of well annotated medical image data available to researchers. One of the first things researchers are told is that we require significant expertise to reliably and accurately interpret and label such data. We see significant inter- and intra-observer variability between expert annotations of medical images. Still, it is a widely held assumption that novice annotators are unable to provide useful annotations for use by clinical Deep Learning models. In this work we challenge this assumption and examine the implications of using a minimally trained novice labelling workforce to acquire annotations for a complex medical image dataset. 
We study the time and cost implications of using novice annotators, the raw performance of novice annotators compared to gold-standard expert annotators, and the downstream effects on a trained Deep Learning segmentation model's performance for detecting a specific congenital heart disease (hypoplastic left heart syndrome) in fetal ultrasound imaging. %We apply a 2D ultrasound dataset of 4CH fetal heart views, and ask annotators for a pixel-level segmentation of 4 chambers and the whole heart. We show that...
%%%
\end{abstract}

\section{Introduction}

It is commonly believed that domain experts are the only reliable source for annotating medical image data. This assumption has resulted in a dearth of annotated medical image datasets due to the time and high costs associated with expert labelling time. In this work we challenge this assumption and employ novice annotators to perform a complex multi-class fetal cardiac ultrasound (US) segmentation task. 

A core goal of medical image analysis is to free up experts' time for more challenging tasks and time with patients. Our current view is that expert annotation efforts that aid in the development of models, will save expert time in the long term. However we hypothesise that in many cases, this annotation effort can be performed by novice annotators at a lower cost, saving both resources and experts' time, with minimal impact on the performance of automated downstream models.

%We compare and analyse the quality of novice annotations, and the quality of models trained using novice annotations to challenge the assumption that experts are the only reliable annotators for medical image data. 

Segmentation is widely regarded as among the most labour intensive medical image analysis tasks, requiring pixel-level labels to enable supervised learning methods to learn complex segmentation tasks. In this study we use a multi-class fetal cardiac US segmentation task as our initial test case, as this task is challenging in both anatomy and modality (noisy, heterogeneous and often contains artefacts). This makes the task of annotating fetal US images challenging for both experts and novices, and an ideal test case for comparing the efficacy of novice annotations. Segmentation of the fetal heart from '4-Chamber view' images provides quantitative biomarkers that can be used for the diagnosis of Hypoplastic Left Heart Syndrome (HLHS). As such we include in our dataset several HLHS cases. The presence of pathology within our dataset makes this annotation task even more challenging, and enables us to compare the performance of novice and expert annotations on a segmentation-informed diagnostic classification task.% as well as segmentation. \\

We provide evidence that the reliability of novice annotators is greater than expected and that this approach might be a viable option for annotation of medical image datasets in the future.

\noindent\textbf{Related work:}
Significant work has been done to mitigate for a lack of well-annotated medical imaging data. Learning from fewer labels, unsupervised learning and active learning are all valuable contributions in this and their benefits go beyond our setting. Advances in these fields can only benefit from the increasing sizes of annotated medical image datasets, and as such we do not challenge these approaches. More tightly related to our work are methods for learning from crowd-sourced noisy labels, where annotations of varying quality are acquired~\cite{revolt,Yu2020RobustnessSegmentation,Tajbakhsh2020EmbracingSegmentation,Tinati2017AnProject,RodriguesDeepCrowds,Cheplygina2016EarlyCT}.

In \cite{snow-etal-2008-cheap} it is shown that novice annotators are comparable to expert annotators for a series of natural language annotation tasks, and that only a small number of novice annotations are necessary to equal the performance of expert annotators. In \cite{fang-etal-2010-pruning} it is shown that novice annotators are able to effectively prune non-informative text from training data for sentiment classifiers to improve classification performance of trained models.

In \cite{jamison-gurevych-2014-needle} it is shown that crowd-sourcing many noisy labels for heavily class imbalanced text classification datasets is expensive and the usual benefits of redundant labelling seen in crowd-sourcing scenarios is lesser in imbalanced settings. \cite{jamison-gurevych-2014-needle} provide techniques for discarding redundant instances such that annotations can be acquired in a cost-effective way over a five-way majority vote aggregation.

In \cite{Wilm2020HowAssessment} the authors assess the effects of aggregating progressively more labels per instance on model performance for mitotic figure detection from histologic images. They show that high accuracy can be achieved with a single annotation per image, and improved by aggregating three annotations per image, while aggregating beyond three annotations per image results in only minor very minor performance increases.

% An alternative to providing detailed annotation instructions for crowd-sourcing is proposed in \cite{revolt} for image level classification tasks. The idea is to collect disagreements between annotators to identify ambiguous concepts and to find groups of semantically related items for post-hoc label decisions. They show that their method required an upfront cost but this is mitigated by the re-usability of learnt annotation structures without requiring new data to be collected.

In \cite{crowd-quality} criteria are proposed by which the suitability of a text sentiment classification task for crowdsourcing can be assessed (1. Noise level, 2. Inherent Ambiguity and 3. Informativeness to the model). Models trained on expert and novice annotations are compared. By considering the three proposed criteria, it is shown that comparable model performance can be achieved using expert or novice annotations. 

% In \cite{Yu2020RobustnessSegmentation} the authors show that Deep Learning models currently used for medical image segmentation can be robust to noisy training data. They demonstrate that no significant difference in model performance is seen when using up to 20\% noisy cases for training on a CT mandible segmentation challenge.

For a 3d segmentation correction task, there is evidence for little to no difference between novices and expert performance (engineers with domain knowledge, medical students, and radiologists) in the ability to detect and correct errors made by a segmentation algorithm \cite{Heim2018Large-scaleAlgorithms}, although novice annotators need significantly more time per annotation. \\

\begin{figure}
    \centering
    \includegraphics[width=\textwidth]{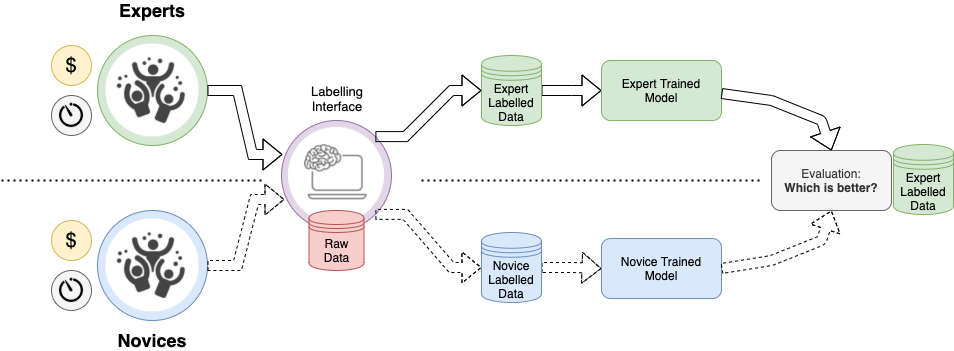}
    \caption{Graphical overview of our process: We assess the upstream and downstream impacts of using novice annotations in place of expert annotations on a challenging medical image segmentation task. Each set of annotations is acquired, pre-processed and used to train models in the same way. We evaluate both expert and novice models against an expert annotated test set.}
    \label{fig:overview}
\end{figure}

\noindent\textbf{Contribution:}
We assess the upstream and downstream impacts of training medical image multi-class segmentation models, and downstream classification models on noisy labels from novice annotators compared against gold-standard labels from expert annotators. 

We show that novice annotators are capable of performing complex medical image annotation tasks to a high standard, and that variability between novices and experts is comparable to that amongst experts themselves. We show that models trained on novice labels are comparable to those trained on expert labels for multi-class segmentation and downstream classification.

We analyse the time and costs associated with using expert vs. novice labels to show that using novice annotations is more resource efficient, and that the major parameter governing model performance is dataset size, rather than label quality in this setting. This will enable clinical and translational researchers to develop a greater understanding of the trade-offs associated with acquiring medical image annotations with respect to cost, time and supervised learning method performance.\\

\section{Method}

\noindent\textbf{Annotation Labels collection:} The current paradigm for collecting annotations for medical image data is to present experts with un-annotated data in an annotation interface that allows them to delineate structures of interest in every image (Figure \ref{fig:overview}). Once complete, the annotations and input can be exported for use. In this work we employ novice annotators to perform the same task using the same annotation tools on the same data to provide us with novice annotated for later use, as shown in the bottom half on Figure \ref{fig:overview}. We use the Labelbox web-based interface as our annotation tool \cite{Labelbox}.\\

\begin{figure}[ht]
     \centering
     \begin{subfigure}[b]{0.24\textwidth}
         \centering
         \includegraphics[width=\textwidth]{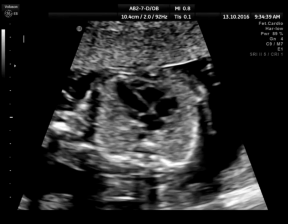}
     \end{subfigure}
     \begin{subfigure}[b]{0.24\textwidth}
         \centering
         \includegraphics[width=\textwidth]{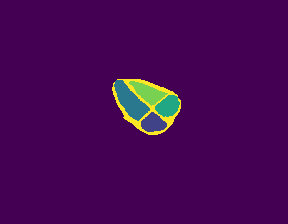}
     \end{subfigure}
     \begin{subfigure}[b]{0.24\textwidth}
         \centering
         \includegraphics[width=\textwidth]{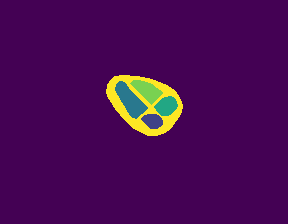}
     \end{subfigure}
     \vfill
     \begin{subfigure}[b]{0.24\textwidth}
         \centering
         \includegraphics[width=\textwidth]{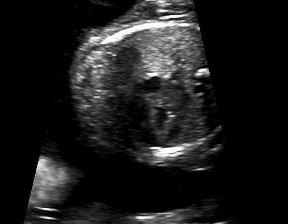}
     \end{subfigure}
     \begin{subfigure}[b]{0.24\textwidth}
         \centering
         \includegraphics[width=\textwidth]{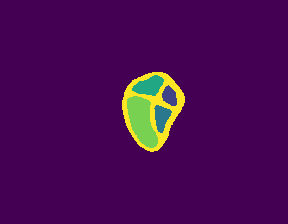}
     \end{subfigure}
     \begin{subfigure}[b]{0.24\textwidth}
         \centering
         \includegraphics[width=\textwidth]{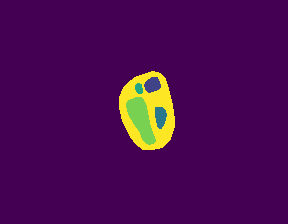}
     \end{subfigure}
        \caption{Example US images and manual segmentations of anatomical areas. Top row: Healthy image, expert manual label and novice manual label (left to right). Bottom row: HLHS image, expert manual label and novice manual label (left to right)}
        \label{fig:examples}
\end{figure}

\noindent\textbf{Segmentation model:} From a single US image of the `4-Chamber Heart View' (4CH view) acquired during fetal screening, we train a model to delineate 5 anatomical areas: \textit{`Whole Heart'} (WH), \textit{`Left Ventricle'} (LV), \textit{`Right Ventricle'} (RV), \textit{`Left Atrium'} (LA) and \textit{`Right Atrium'} (RA) (Figure~\ref{fig:examples}).

We use the UNet architecture as our segmentation network~\cite{RonnebergerU-Net:Segmentation}, known to perform well for US segmentation. We train using dropout~\cite{Gal2016DropoutGhahramani}, for a fixed number of epochs, then select the best performing model on the validation set. Random horizontal and vertical flipping, cropping, translation, rotation and scaling is applied during training. \\

\noindent\textbf{Classification model:} We extract numerical features from $\hat{y_i}^{seg}$ (manual or automated segmentation) in order to classify HLHS vs. healthy patients from interpretable features $f = \{f_0, f_1, ..., f_N\}$ where $f_i = r_{ab} = A_a / A_b$ if $a \neq b$ and $r_{ba}$ is not in $f$ already. Here $r_{ab}$ is the ratio between two quantities and consider $r_{ab}$ and $r_{ba}$ to contain equivalent information and exclude the latter from $f$. $A_a$ is the count of pixels belonging to class $a$ in $\hat{y_i}^{seg}$ which acts as an estimate to the area.

We apply an L2 regularised, class weight balanced logistic regression classifier implementation to classify the extracted segmentation area ratio features as healthy vs. HLHS as in~\cite{Budd2021DetectingMaps}.\\

\noindent\textbf{Statistical analysis:} Here we pose the questions answered in this paper and outline our approach to answering them. For tests of statistically significant difference between distributions we use a two-tailed Z-test for the null hypothesis of identical means: $ Z = \frac{\hat{X} - \mu_0}{s} $ where $Z$ is our test statistic, $\mu$ is our population mean and $s$ is our population standard deviation.
\\

\textbf{Q1: Are novice annotations as similar to experts as expert annotations are to other experts?} We answer this question by computing the average DICE similarity coefficient between novice  and expert annotations, and between pairs of expert annotations. We calculate the Dice score for each class separately, and test for statistically significant difference between the two sets.\\

\textbf{Q2: How different are automated segmentations trained on experts annotations to automated segmentations trained on novice annotations?} We show evidence by computing the average DICE similarity coefficient between novice and expert trained model predictions and an expert annotated test set. We calculate the DICE score for each class separately, and test for statistically significant difference between the two sets.\\

\textbf{Q3: How different are classification predictions trained on either manual, or model based segmentations from novices compared to experts?} We train a classifier using training data from manual expert and manual novice annotations, as well as expert model and novice model predictions. We test each classifier on our expert test set and compare key performance metrics to assess the discrimitive powers of novice vs. expert based segmentations.\\

\textbf{Q4: In resource limited scenarios are expert or novice annotations more cost effective to attain the same model performance?} We observe the time/cost/quality trade-off by measuring the DICE scores obtained by models trained using novice and expert data on progressively more labels (50 to 1000 labels) using a UNet with 200 epochs. We use DICE scores on the test set as a measure of prediction quality, and use time taken and estimated financial cost to acquire each annotation, to plot the time vs. cost vs. performance of our models for both experts and novice annotations.\\

\section{Experiments and Results}

\noindent\textbf{Data and Pre-processing}:

\begin{itemize}
    \item Raw images: We use a private and ethics/IP-restricted, de-identified dataset of 2380 4CH US images, with 1000 for training, 380 for validation, 1000 for testing acquired on Toshiba Aplio i700, i800 and Philips EPIQ V7 G devices.
    \item Expert segmentations: A fetal cardiologist and three expert sonographers delineated the images using Labelbox~\cite{Labelbox}. Multiple expert annotations for 319 images were acquired and used to calculate expert-expert annotation similarity. A single annotation for each image is used for training.
    \item Non-expert segmentations: A novice workforce with no experience annotating medical US data was employed to delineate the images using Labelbox~\cite{Labelbox}, this workforce was provided with the instruction pdf included in the Supplementary Material. Three novice annotations for every image in the training set were acquired, each further calculation made used novice annotations was performed three times and the results averaged.
    \item Time: Experts annotated images in an average time of 127s per image, and Novices annotated images in an average time of 253s per image.
    \item Cost: Experts costs were set at \$60 per labelling hour, and Novices cost \$6 per labelling hour.
\end{itemize}

During analysis 10 cases were found to have two hearts visible (split screen view), resulting in zero DICE agreement amongst experts and experts and novices, having annotating different sides of the image. These cases have been removed. A significant proportion of the worst performing remaining cases are a result of mislabelling of left/right atriums and ventricles resulting in very low DICE scores for those cases. \\

\begin{figure}[]
    \centering
     \begin{subfigure}[b]{0.49\textwidth}
         \centering
         \includegraphics[width=\textwidth]{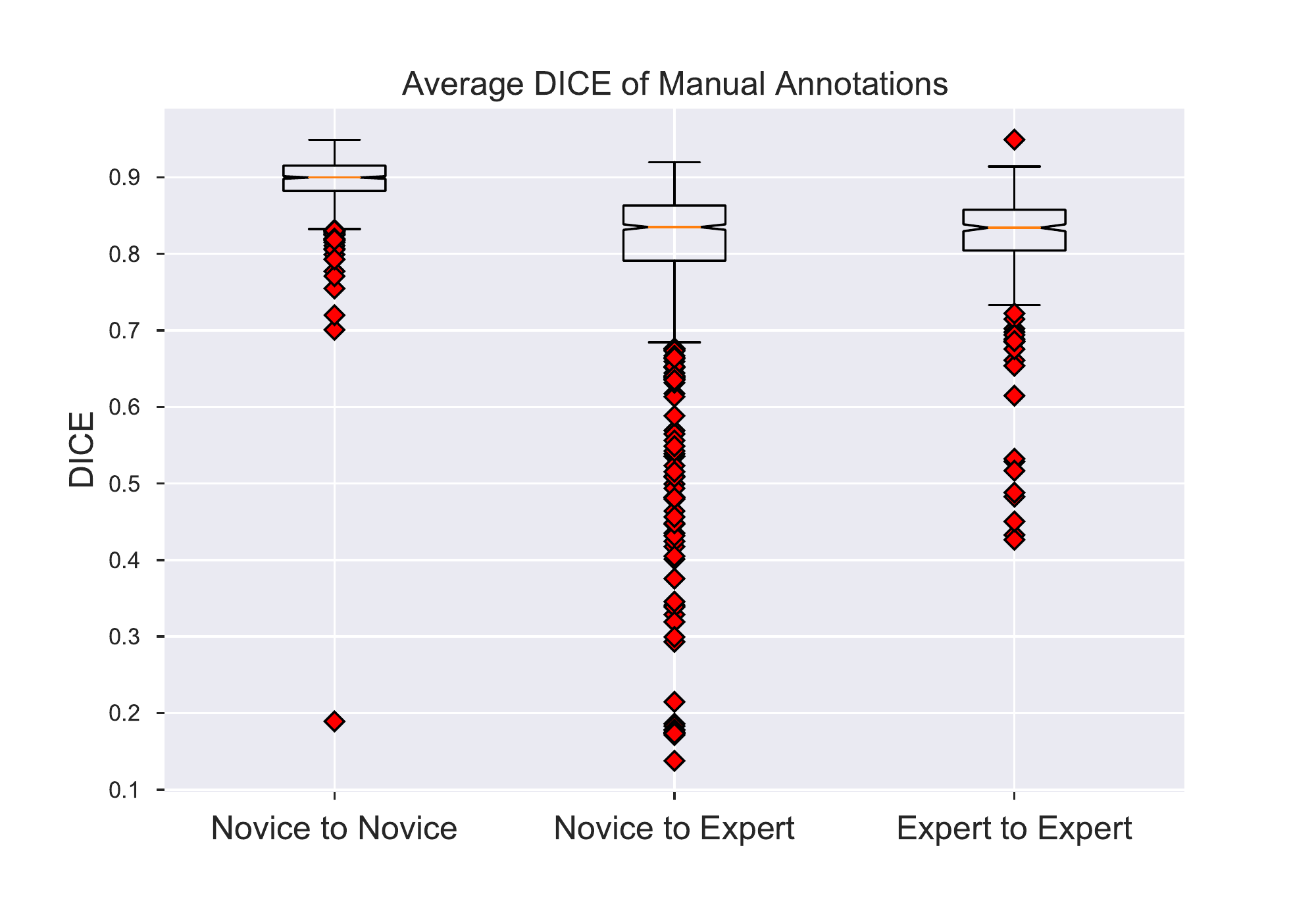}
     \end{subfigure}
     \hfill
     \begin{subfigure}[b]{0.49\textwidth}
         \centering
         \includegraphics[width=\textwidth]{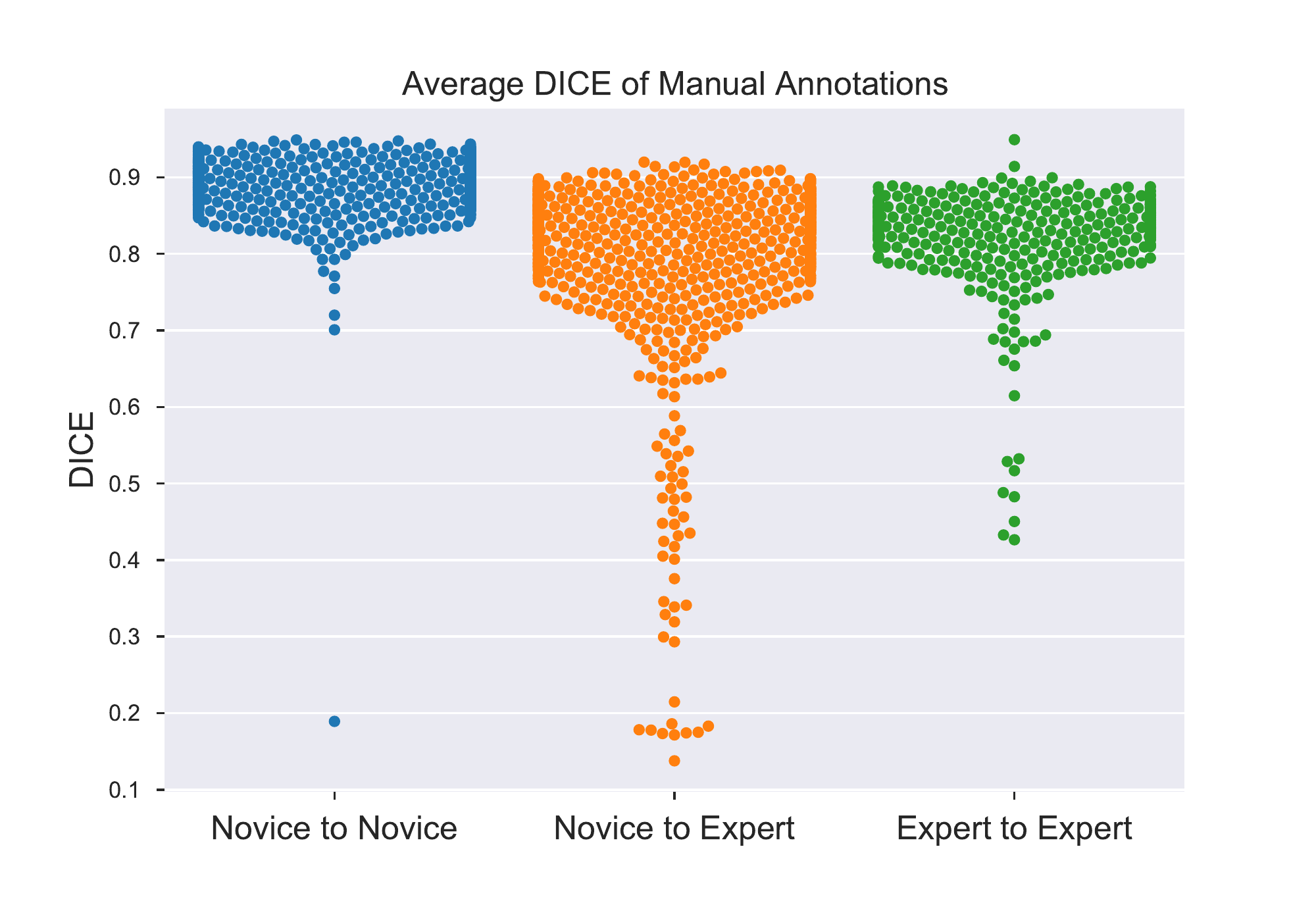}
     \end{subfigure}
    \caption{Distributions of DICE similarity scores between raw labels for Novice to Novice labels, Novice to Expert labels and Expert to Expert Labels. Left: Box and whisker plots. Right: Swarm plots.}
    \label{fig:dice:raw}
\end{figure}

\noindent\textbf{Q1:} In Table \ref{dice:raw} the average DICE scores for expert-expert and novice-expert segmentations show that no statistical difference is found between the variability of annotators on three out of five annotated classes. This shows that novice annotators are better at annotating complex medical data than is assumed and the variability between experts and novices is similar to that amongst experts for these three classes. Figure \ref{fig:dice:raw} highlights the similarity in DICE distributions between novice-novice and expert-novice annotations, indicating it may not be possible to avoid variation in annotations even when using experts annotations alone.\\

\begin{table}[ht]
\centering
\begin{tabular}{@{}lcccccc@{}}
\toprule
 DICE                       & &LV     & RV     & LA    & RA & WH \\ \midrule
Expert to Expert & &0.807 & 0.787 & 0.764 & 0.808 & 0.887 \\
Novice to Expert & &0.778 & 0.761 & 0.757 & 0.806 & 0.894 \\
p-value                & &\textbf{0.009} & \textbf{0.005} & 0.551 & 0.866 & 0.359 \\ \bottomrule
\end{tabular}
\caption{Mean DICE scores of manual annotations performed by Experts compared with DICE scores of manual annotation performed by Novices. Statistically significant (95\%) results shown in bold.}\label{dice:raw}
\end{table}

\noindent\textbf{Q2:} Average DICE scores and segmented class sizes for expert trained vs. novice trained models show there is no statistical difference in the performance of the models on three out of five classes (Tables \ref{dice:model}, and \ref{size:model} (Supplementary Material)). Expert models average higher DICE scores on all but one class, and one reason for this better performance is that both models are tested against an expert annotated test set. Models trained on novice annotations can perform almost equally well as those trained on expert annotations for multi-class US segmentations problems. We see a significant difference between average class sizes predicted by the two models, most noticeably in the right ventricle class (RV), however their overall similarity is highlighted in Figures \ref{fig:model} and Figures \ref{fig:model:size}-\ref{fig:model_class} in the Supplementary Material where both DICE and sizes appear very similar across all classes.\\

\begin{figure}[h]
     \centering
     \begin{subfigure}[b]{0.49\textwidth}
         \centering
         \includegraphics[width=\textwidth]{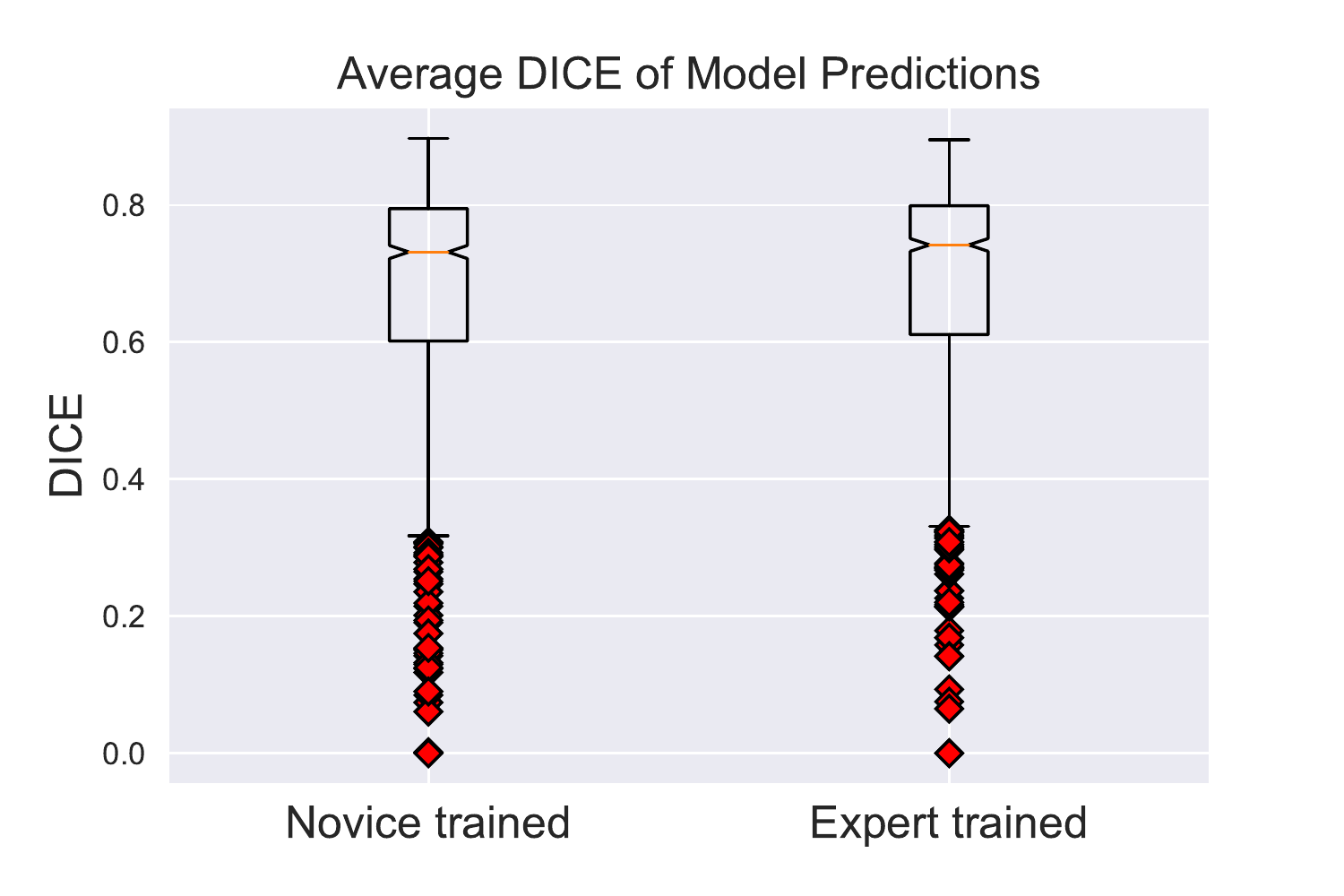}
     \end{subfigure}
     \hfill
     \begin{subfigure}[b]{0.49\textwidth}
         \centering
         \includegraphics[width=\textwidth]{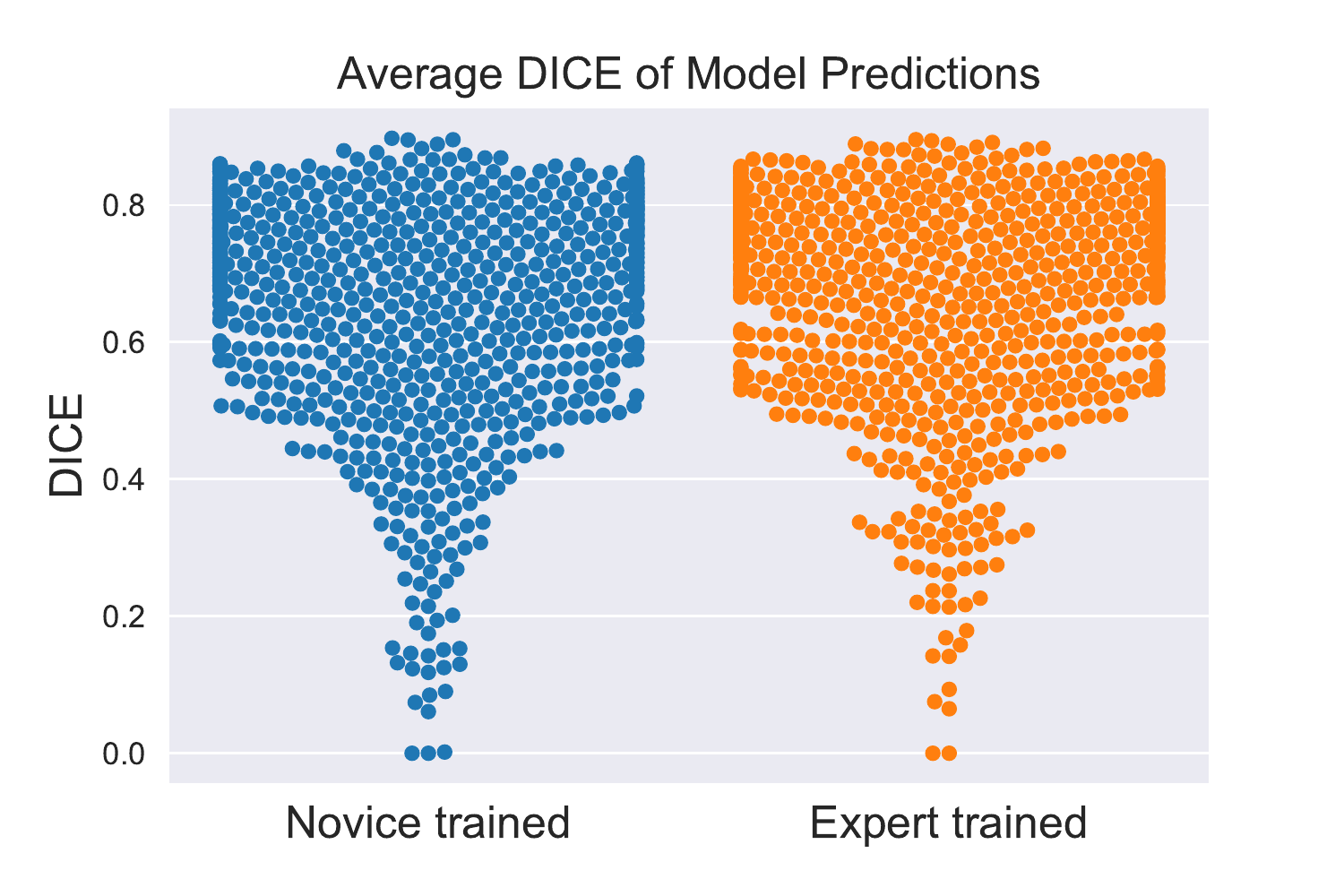}
     \end{subfigure}
        \caption{Distributions of segmentation predictions DICE scores against the expert test set}
        \label{fig:model}
\end{figure}

\begin{table}[ht]
\centering
\begin{tabular}{@{}llrrrrr@{}}
\toprule
DICE               &  & LV     & RV       & LA     & RA     & WH     \\ \midrule
Expert Model &  & 0.721  & 0.707    & 0.663  & 0.749  & 0.617  \\
Novice Model &  & 0.708  & 0.679    & 0.652  & 0.731  & 0.634  \\
p-value      &  & 0.174  & \textbf{0.003}    & 0.321  & 0.071  & \textbf{0.001} \\
\bottomrule
\end{tabular}
\caption{Class average DICE scores of model predictions and class average sizes in pixels of model predictions, comparing models trained using expert annotations and models trained using novice annotations. Statistically significant (95\%) results shown in bold.}
\label{dice:model}
\end{table}

\begin{figure}[]
     \centering
     \begin{subfigure}[b]{0.49\textwidth}
         \centering
         \includegraphics[width=\textwidth]{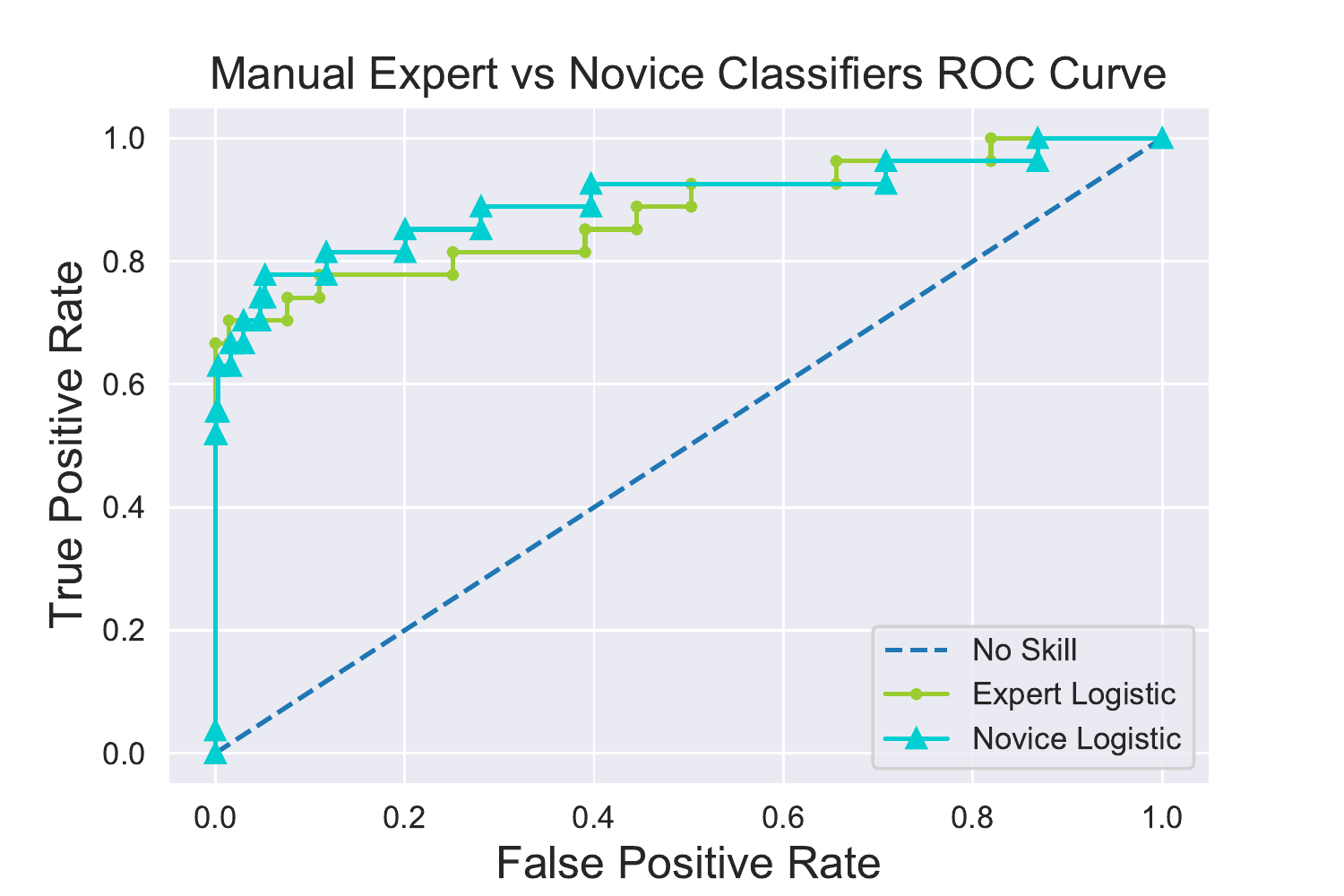}
     \end{subfigure}
     \hfill
     \begin{subfigure}[b]{0.49\textwidth}
         \centering
         \includegraphics[width=\textwidth]{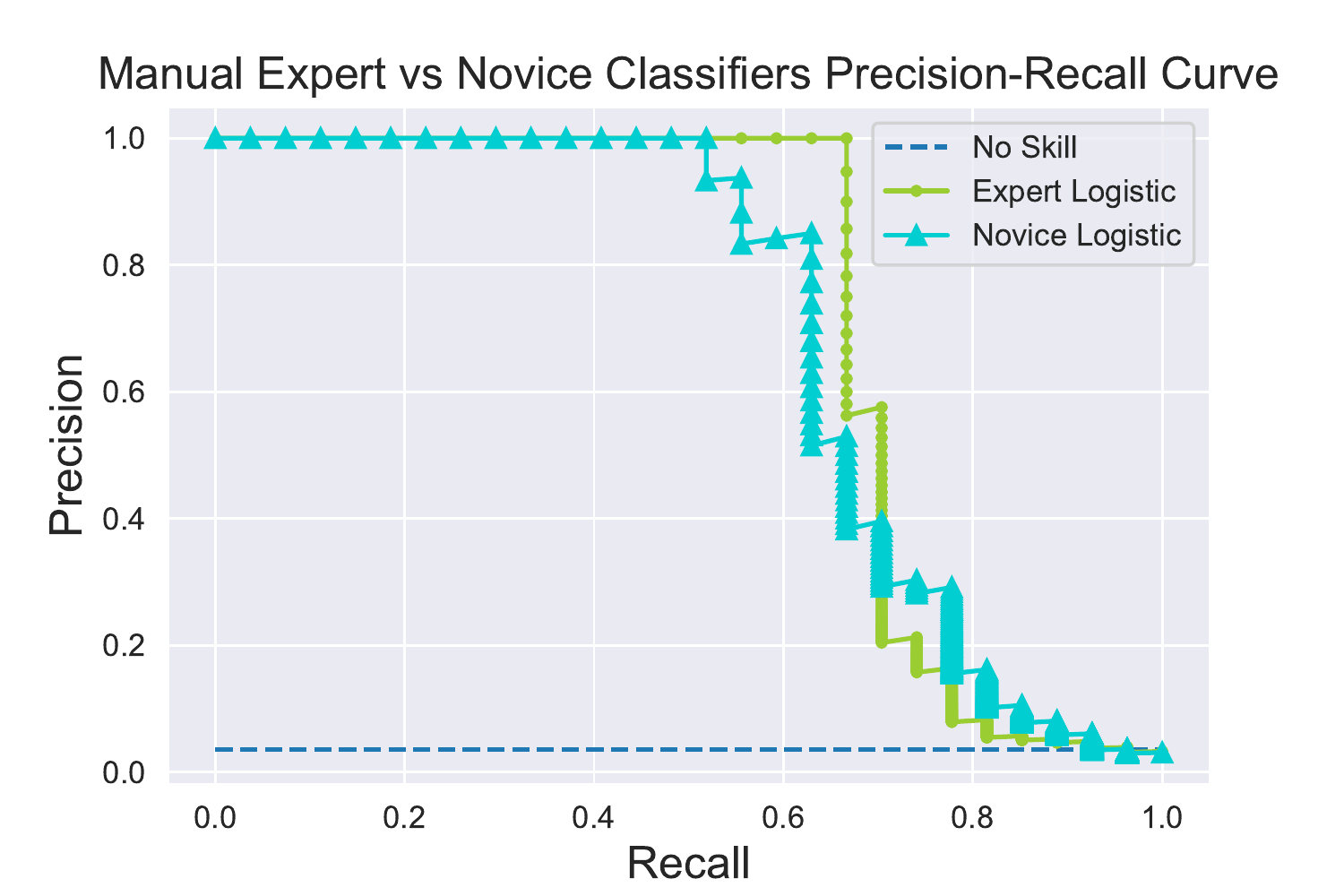}
     \end{subfigure}
     \vfill
     \begin{subfigure}[b]{0.49\textwidth}
         \centering
         \includegraphics[width=\textwidth]{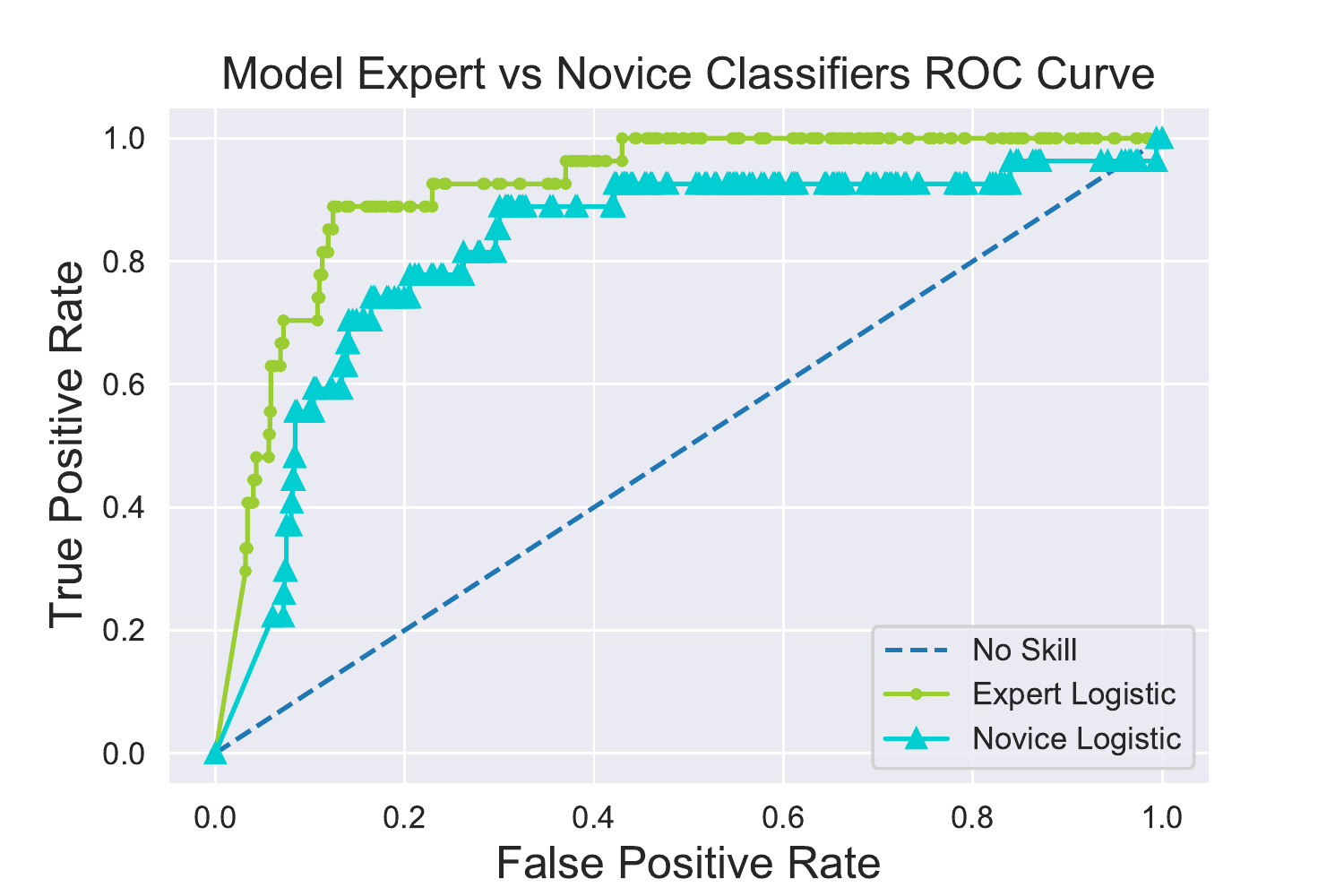}
     \end{subfigure}
     \hfill
     \begin{subfigure}[b]{0.49\textwidth}
         \centering
         \includegraphics[width=\textwidth]{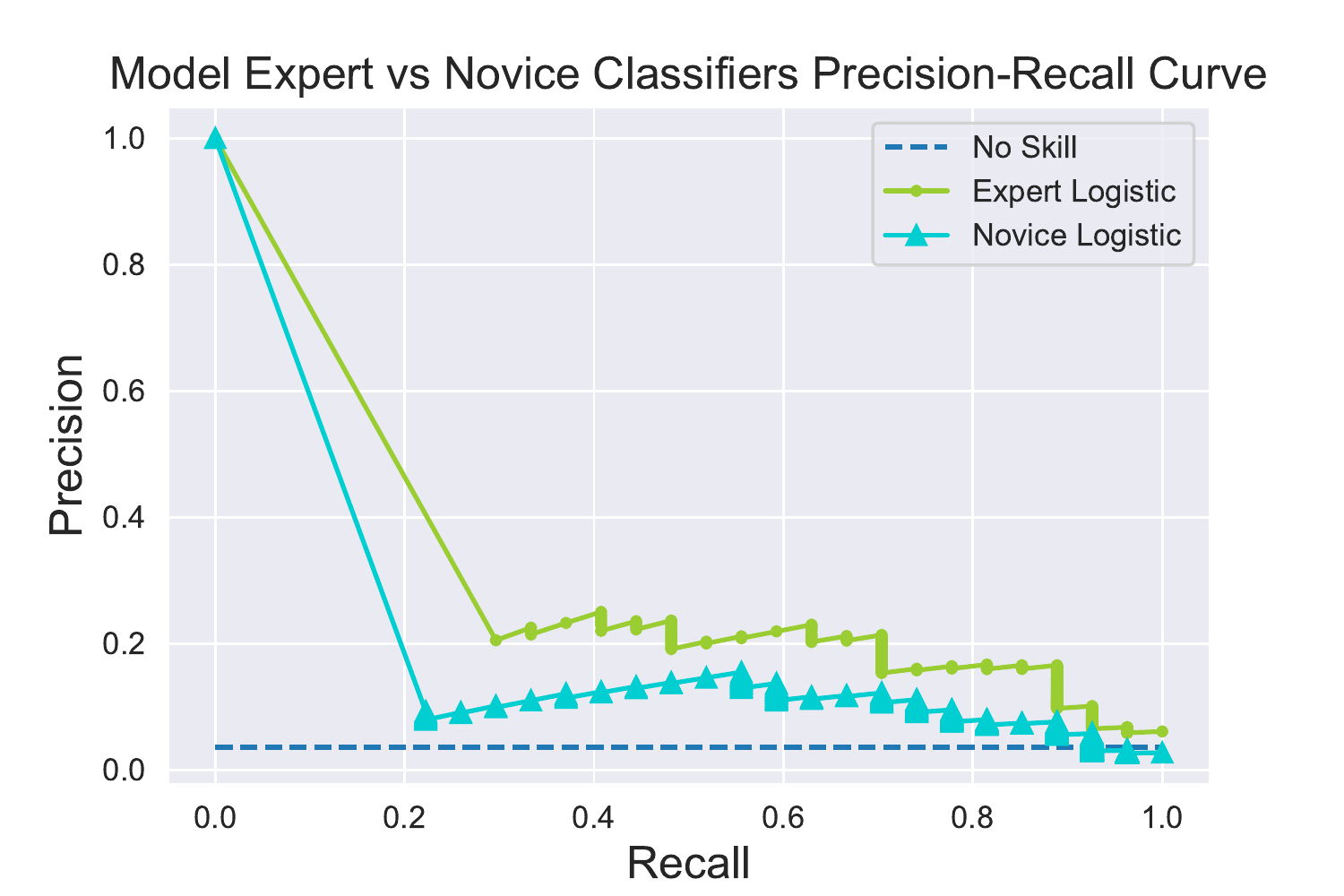}
     \end{subfigure}
        \caption{Top row: Classification performance for manual annotations predicted classifications. Bottom row: Classification performance for model predicted classifications. Left to right: ROC Curves and Precision-Recall curves.}
        \label{fig:manual_curves}
\end{figure}

\noindent\textbf{Q3:} The results of HLHS classification methods trained on manual and automated expert and novice segmentations show that novice trained models attain very similar results to those trained by experts, in both manual and model cases (Table \ref{classification}). We see a slight improvement from expert annotations in Precision and F1 scores but the overall performance is remarkably similar. This result again shows the viability of acquiring a significant proportion of medical image annotations from non-experts during annotation efforts. Figure \ref{fig:manual_curves} highlights that in some scenarios novice manual annotations may out-perform expert annotations on some metrics. Both ROC curves and Precision-Recall Curves for experts and novices follow very similar trajectories demonstrating the similarity in their performance for classification. \\

\begin{table}[ht]
\centering
\begin{tabular}{@{}lrrrrr@{}}
\toprule
          &  & Expert Manual & Novice Manual & Expert Model & Novice Model \\ \midrule
TP        &  & 19            & 20            & 24           & 22           \\
FP        &  & 47            & 64            & 126          & 224          \\
TN        &  & 926           & 909           & 847          & 749          \\
FN        &  & 8             & 7             & 3            & 4            \\
Precision &  & 0.288         & 0.242         & 0.16         & 0.091        \\
Recall    &  & 0.704         & 0.753         & 0.889        & 0.827        \\
F1        &  & 0.409         & 0.367         & 0.271        & 0.165        \\
AUC-ROC   &  & 0.879         & 0.900         & 0.915        & 0.829        \\ 
\bottomrule

% \vfill
\end{tabular}
\caption{Classification results: Precision, Recall and F1 scores are reported for the positive prediction class (HLHS)}\label{classification}
\end{table}

\begin{figure}[ht]
    \centering
    \begin{subfigure}[t]{0.49\textwidth}
        \includegraphics[width=\textwidth]{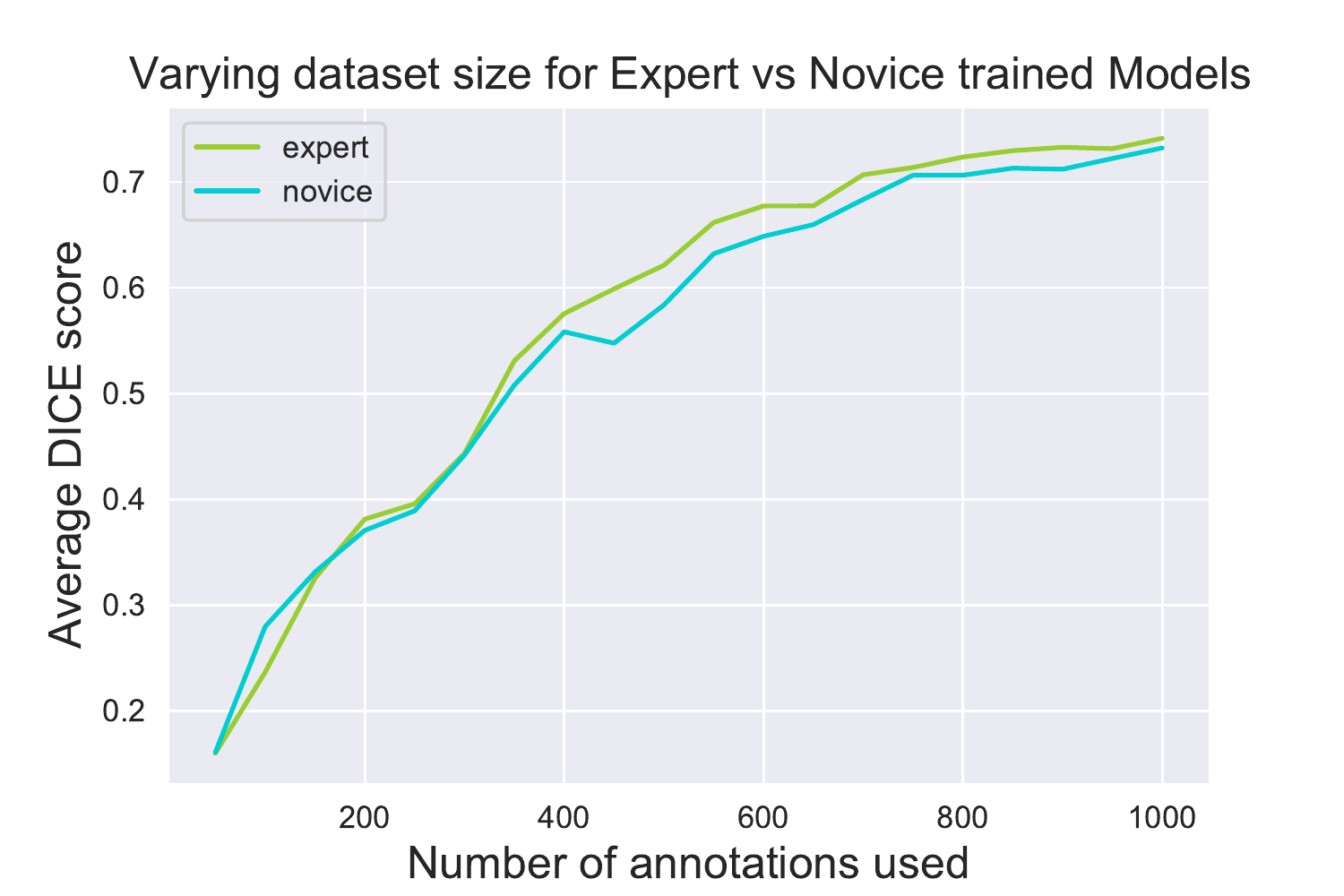}
        \caption{DICE scores as we increase the training dataset size from 50 to 1000 images}
    \label{fig:dice:size}
    \end{subfigure}
    \hfill
    \begin{subfigure}[t]{0.49\textwidth}
        \includegraphics[width=\textwidth]{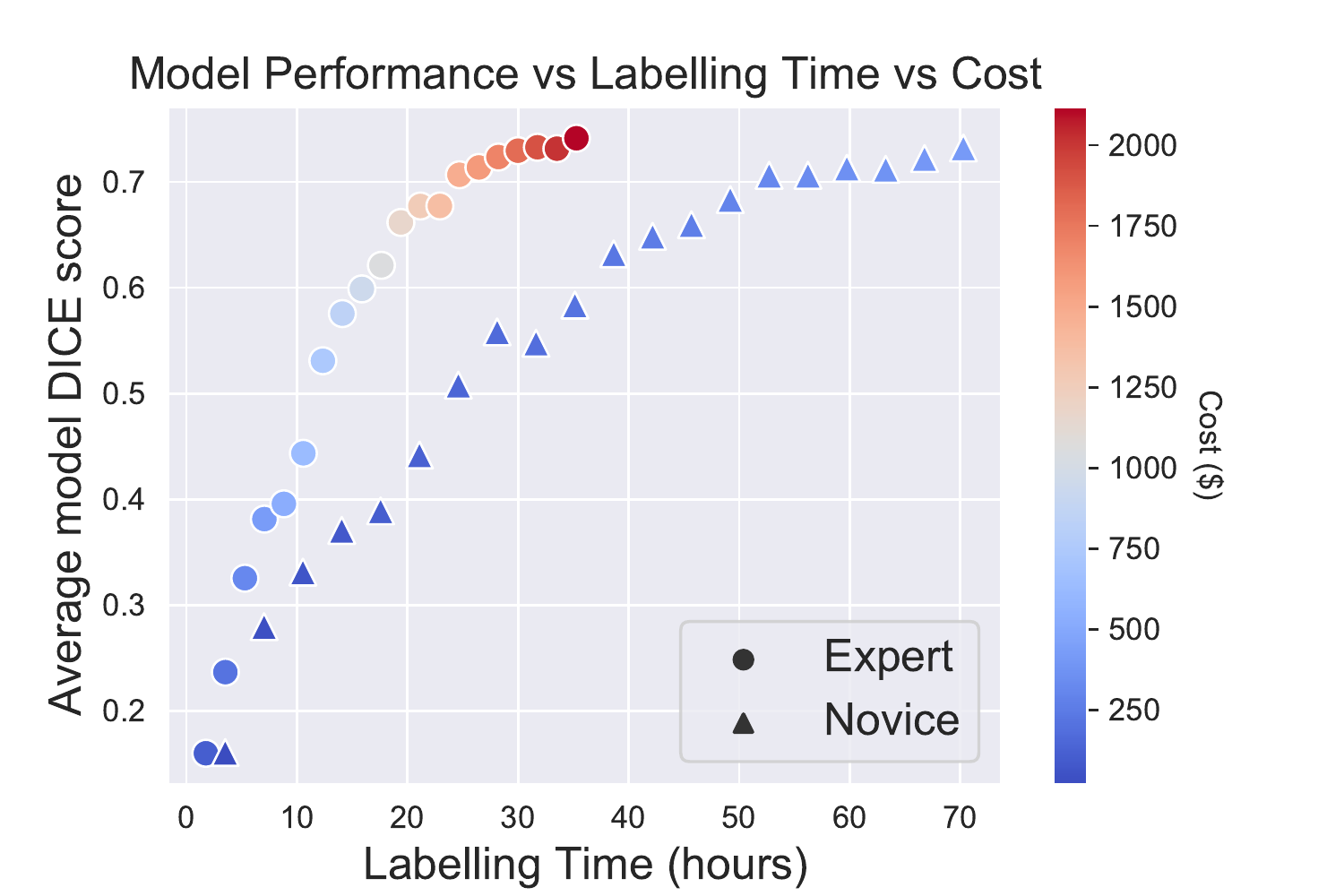}
        \caption{Time, Cost and DICE scores as we increase the training dataset size from 50 to 1000 images}
    \label{fig:cost:size}
    \end{subfigure}
    \caption{Analysis of the Time/Cost/Model performance trade-off.}
\end{figure}

\noindent\textbf{Q4:} Figure \ref{fig:dice:size} shows the consistent increase of both expert and novice trained models as the size of the dataset increases, demonstrating that collecting initial annotations from novices may well suffice to achieve a good accuracy in many tasks. We calculate the cost per image for both novices and experts using the average cost of an hour of labelling work and the average time each annotation took to create. Figure \ref{fig:cost:size} shows how when time is the priority, then expert annotators achieve higher quality models in a shorter time-span, however this comes at a much greater financial cost. If cost is the priority then novice annotators achieve higher quality models at a much smaller financial cost, however the same number of annotations take longer to acquire from novices than from experts. We can see from this that the dominant driving force of improving model quality is dataset size, regardless of whether annotations come from experts or novices, indicating that to train high performing models in a resource efficient way that novice annotations are a useful mechanism by which this can be achieved.\\

\section{Discussion}

We have assessed the upstream and downstream effects of acquiring complex medical image segmentation annotations from novices compared to experts. We have found that raw novice annotations are of remarkable quality, and that novice trained models show only a minor performance decrease compared expert trained models. Our results highlight that annotations performed by novices are of great utility for complex tasks such as segmentation and classification. A time and cost analysis for using limited resources more efficiently is provided, guiding practitioners in acquiring annotations to give the best performing models under their constraints. Through future studies on other complex tasks, we aim to develop protocols through which confidence can be given that novice annotations are sufficient in many use cases.

Additional combination of crowd-sourcing from novice labels with models incorporating measures of annotator skill and merging of multiple annotations show great promise in enabling highly accurate models to be developed on a wide variety of tasks for which expert annotated data has been infeasible to acquire at a large enough scale.

We note that we are unsure of how representative our Labelbox workforce is of the wider novice annotator community. Through our engagement with Labelbox they were made aware of our intentions with the annotated data and it is our hope that no special measures were taken to improve the quality of annotations beyond that of the wider novice annotator community. Similarly, when comparing costs of annotating large datasets, we must consider the ethical implications of employing low-cost workers to perform these tasks - while the low cost makes using workforce services appealing, care must be taken to ensure that workers are paid fairly and under suitable working conditions. Limited information given regarding the locations and working conditions of annotation workforces creates difficultly in making this judgement. Additional consideration must be given to data privacy when using external labelling services, as regulation surrounding data storage and transfer must be adhered to to ensure patient data remains protected. 

\section{Conclusion}

We have demonstrated that novice annotators are capable of performing complex medical image segmentation tasks to a high standard, with a comparable variability to experts as experts show to themselves. We have shown that training models with novice annotations is both resource efficient and can give comparable models in terms of prediction performance against expert annotations for both segmentation and downstream classification tasks. We foresee that in combination with existing methods that better handle noisy annotations, and active learning methods selectively choosing the most informative annotations to acquire next, that novice annotations will play a vital role in developing high-performing models at a fraction of the cost of using expert annotations. 

% \newpage
\bibliographystyle{splncs04}
\bibliography{ref}

\newpage
\section{Supplementary Material}

We provide additional tables and figures as discussed throughout the main text.

\begin{table}[]
\centering
\begin{tabular}{@{}llrrrrr@{}}
\toprule
 Size (px)            &  & LV     & RV       & LA     & RA     & WH     \\ \midrule
Expert Model &  & 806    & 630      & 468    & 612    & 4732   \\
Novice Model &  & 737    & 536      & 428    & 546    & 5130   \\
p-value      &  & \textbf{0.0051} & \textbf{2.45e-07} & \textbf{0.0083} & \textbf{0.0009} & \textbf{0.0018} \\ \bottomrule
\end{tabular}
\caption{Class average sizes in pixels of model predictions, comparing models trained using expert annotations and models trained using novice annotations. Statistically significant (95\%) results shown in bold.}
\label{size:model}
\end{table}

\begin{figure}[h]
     \begin{subfigure}[b]{0.49\textwidth}
         \centering
         \includegraphics[width=\textwidth]{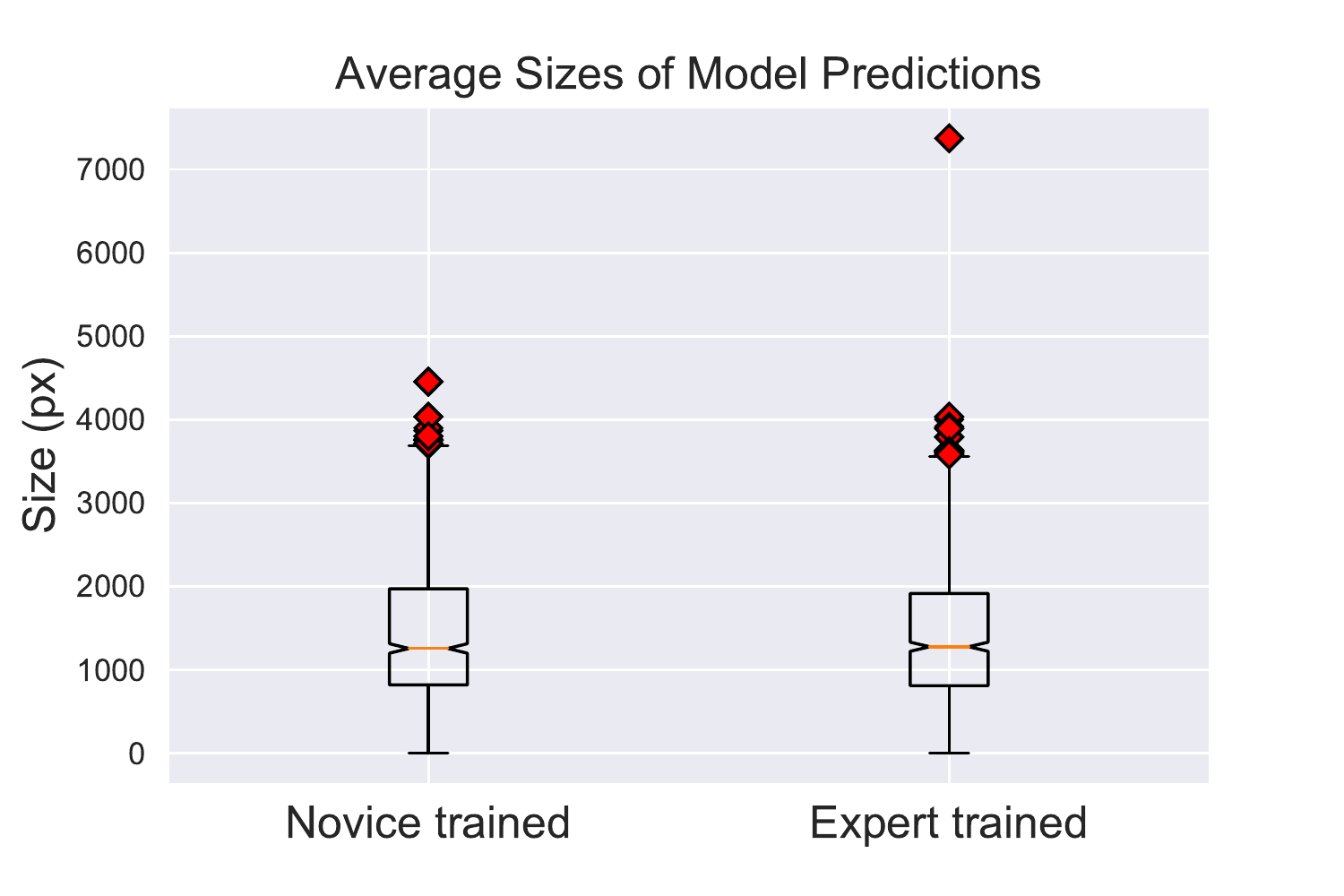}
     \end{subfigure}
     \hfill
     \begin{subfigure}[b]{0.49\textwidth}
         \centering
         \includegraphics[width=\textwidth]{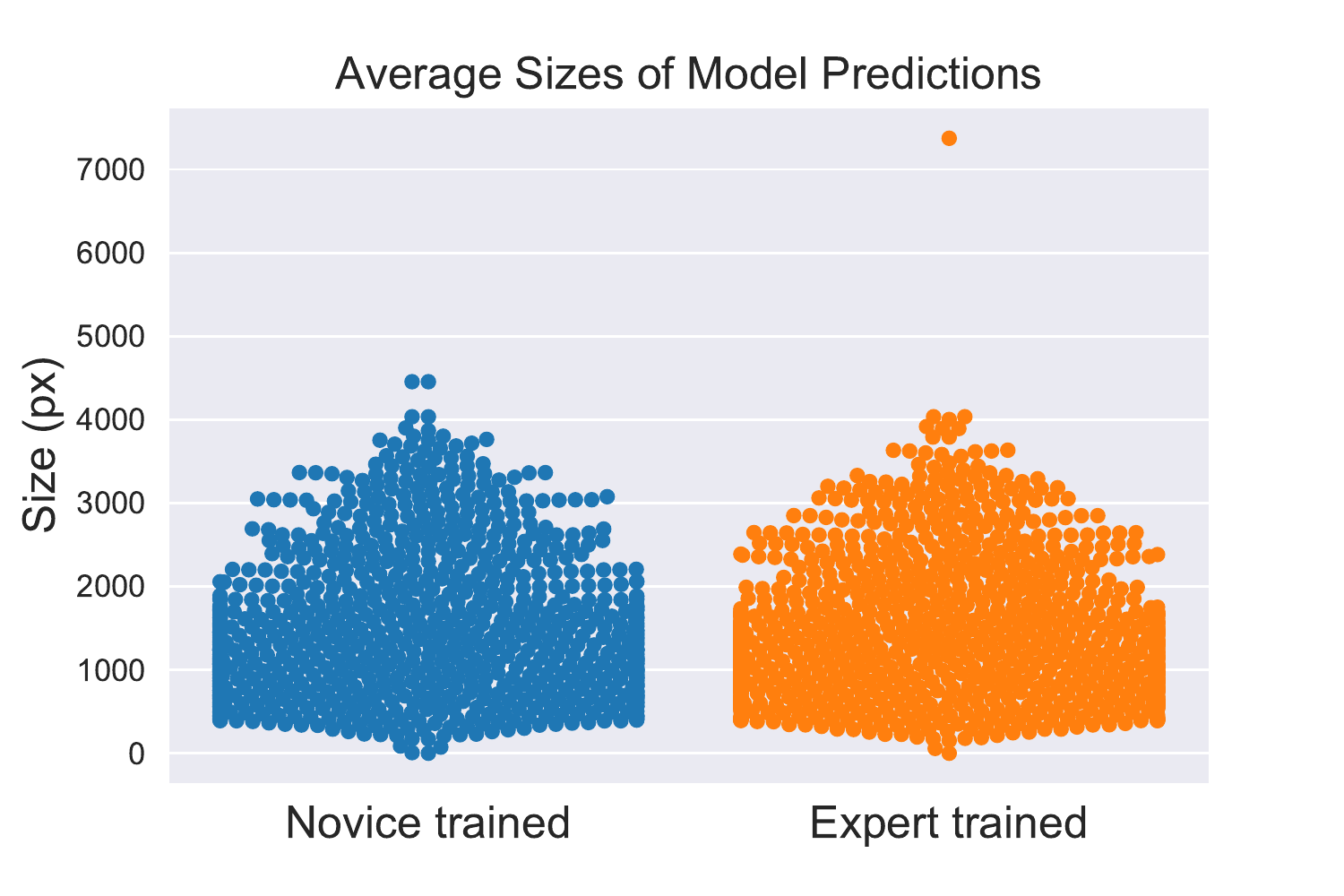}
     \end{subfigure}
        \caption{Distributions of segmentation predictions average pixel sizes.}
        \label{fig:model:size}
\end{figure}

\newpage

\begin{figure}[h]
     \centering
     \begin{subfigure}[b]{0.49\textwidth}
         \centering
         \includegraphics[width=\textwidth]{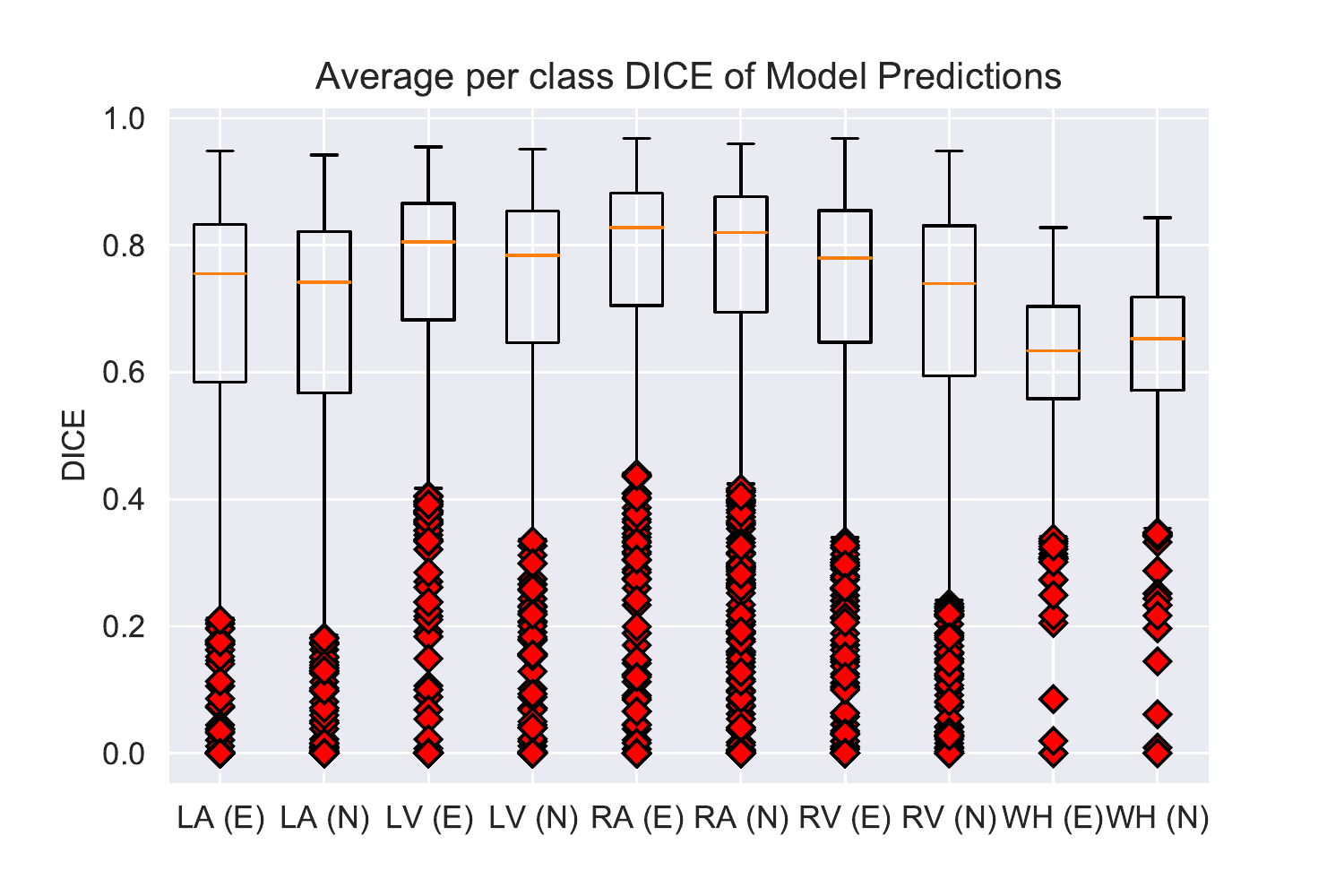}
     \end{subfigure}
     \hfill
     \begin{subfigure}[b]{0.49\textwidth}
         \centering
         \includegraphics[width=\textwidth]{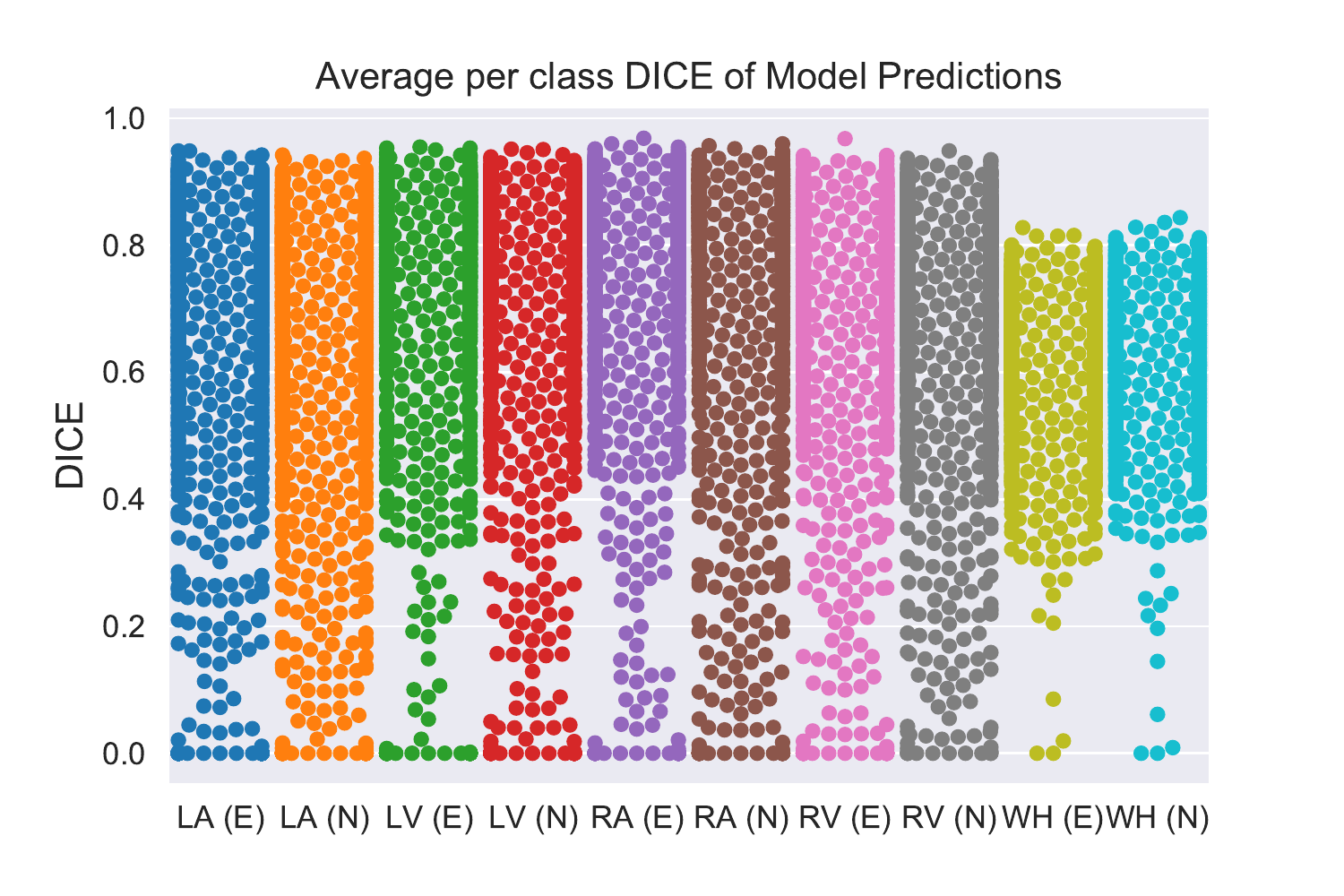}
     \end{subfigure}
     \vfill
     \begin{subfigure}[b]{0.49\textwidth}
         \centering
         \includegraphics[width=\textwidth]{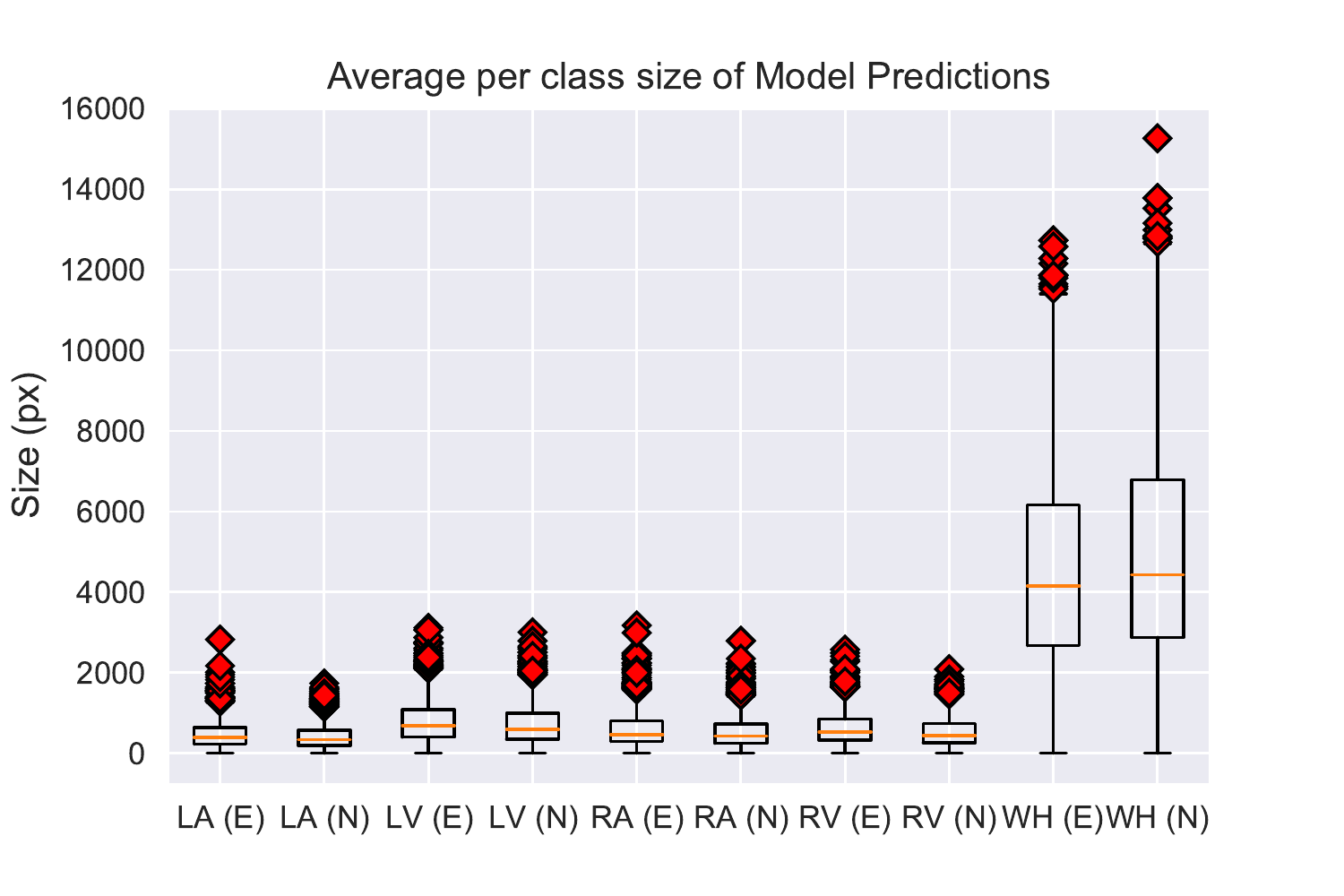}
     \end{subfigure}
     \hfill
     \begin{subfigure}[b]{0.49\textwidth}
         \centering
         \includegraphics[width=\textwidth]{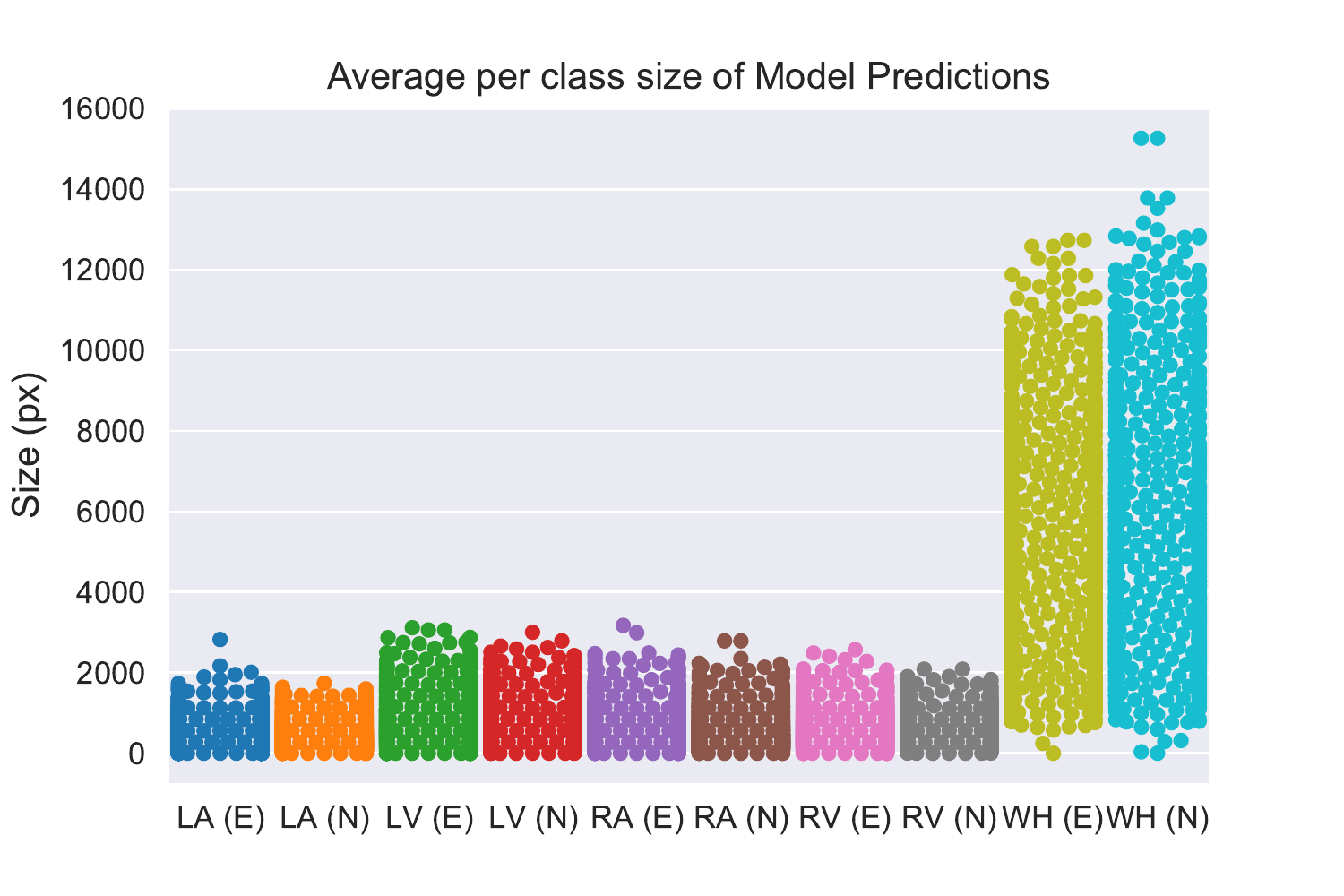}
     \end{subfigure}
        \caption{Top row: Distributions of per class segmentation DICE scores. Bottom row: Distributions of per class segmentation size.}
        \label{fig:model_class}
\end{figure}

\newpage

\newlength{\originalVOffset}
\newlength{\originalHOffset}
\setlength{\originalVOffset}{\voffset}   
\setlength{\originalHOffset}{\hoffset}

\setlength{\voffset}{0cm}
\setlength{\hoffset}{0cm}


\section*{Supplementary material}

## Instructions for Labelbox Workforce

The task at hand is to segment 5 anatomical classes within a fetal ultrasound view (the 4 chamber view), The 5 classes to segment are the 'Whole heart', the 'left ventricle', 'right ventricle', 'left atrium' and 'right atrium'. This is not an easy task, but when done accurately can be of great help in diagnosing congenital heart disease, which is the motivation behind this task.

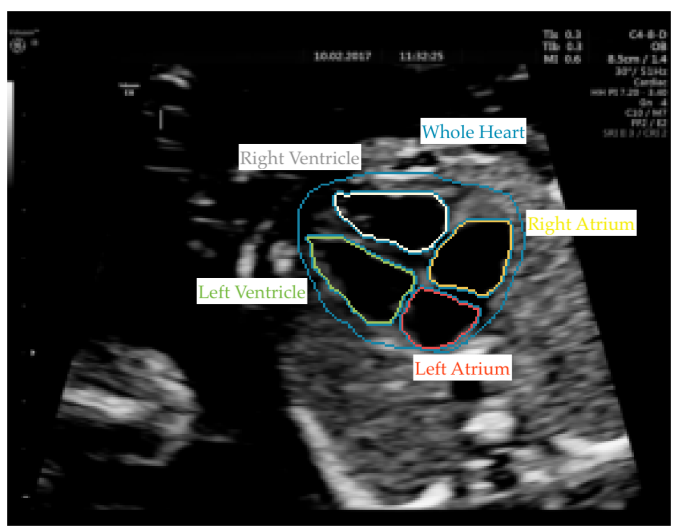

The first class is the 'Whole heart' class that encompasses the four chambers of the heart:

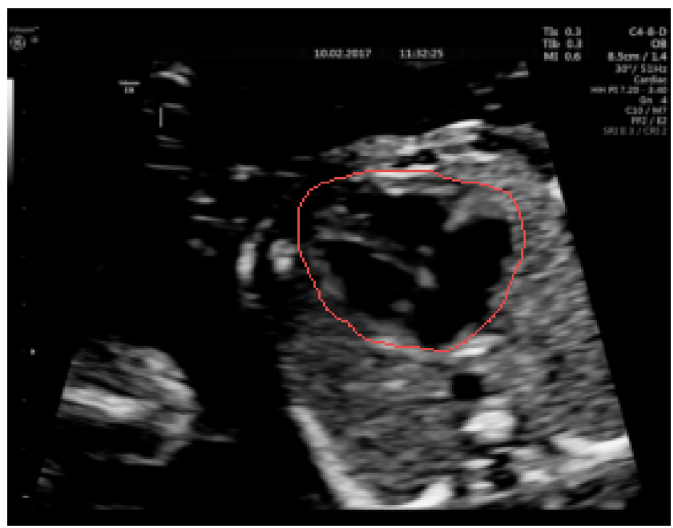

Next is the 'Left Atrium' class:

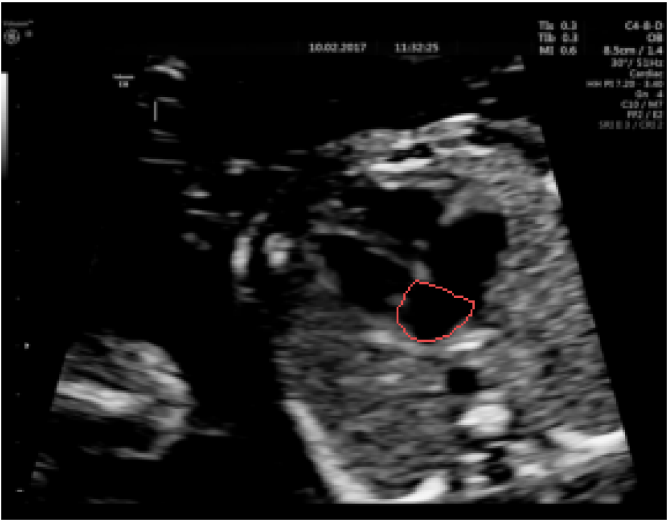

The 'Right Ventricle' class:

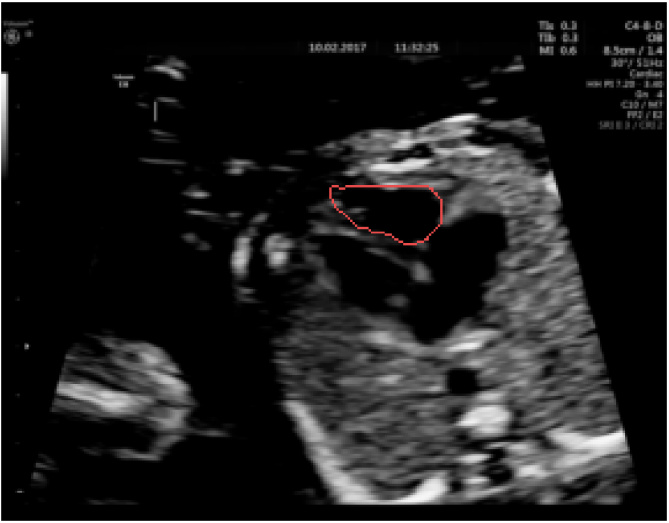

The 'Right Atrium' class:

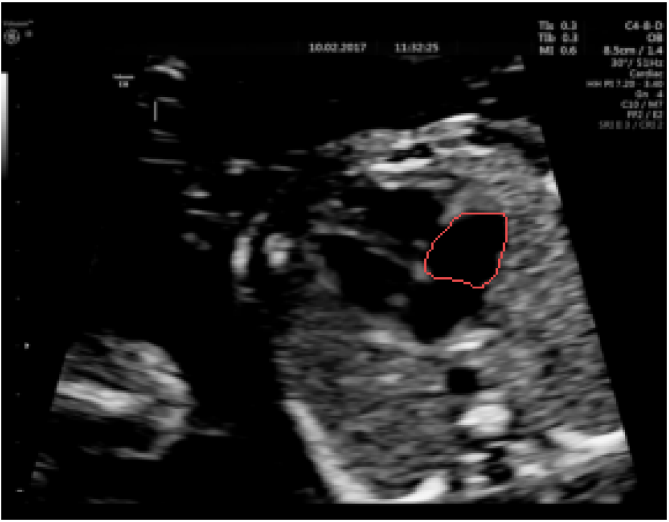

Finally the 'Left Ventricle' class:

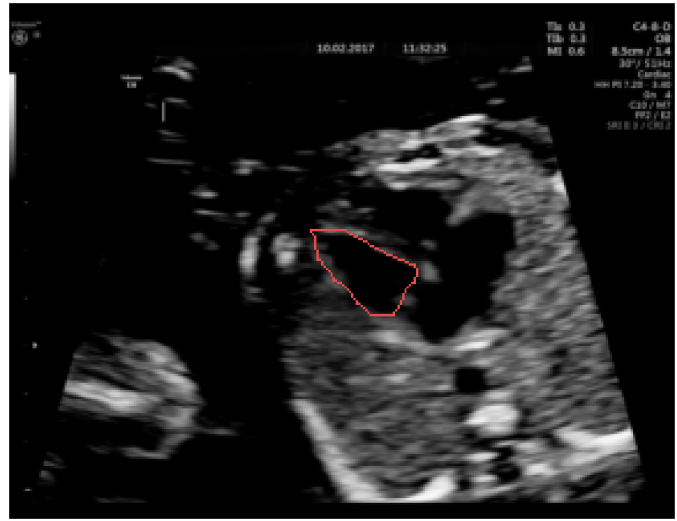

**Challenge 1: Orientation**

When annotating the left/right atrium and ventricles it is important that we correctly identify the orientation of the heart so that the left and right atrium and ventricles can be correctly annotated. This can be done by locating the spine, from which the heart chamber closest to the spine is the 'Left Atrium', and the chamber furthest from the spine is the 'Right Ventricle', and the other two chambers can be annotated with respect to these two chambers. In general for healthy hearts, the ventricles are more elongated than atriums, this rule of thumb can be used to distinguish between atriums and ventricles.

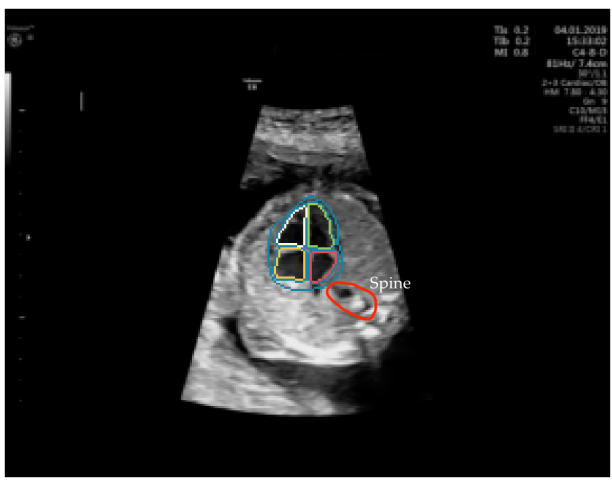
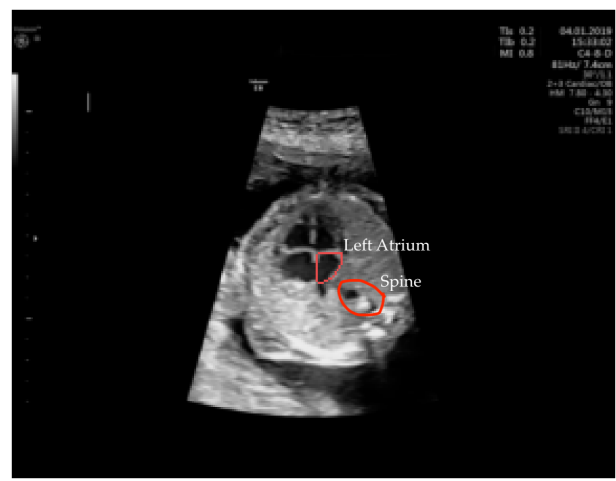

**Challenge 2: Abnormal heart structure**

Not all the hearts in this dataset are healthy, and as such several cases presented to you may exhibit shapes unlike the majority of hearts you will see, however it is important that the true shape of the heart is annotated as it is this difference in shape that can be so important in diagnosis congenital heart disease. We provide some examples of abnormal heart structures below:

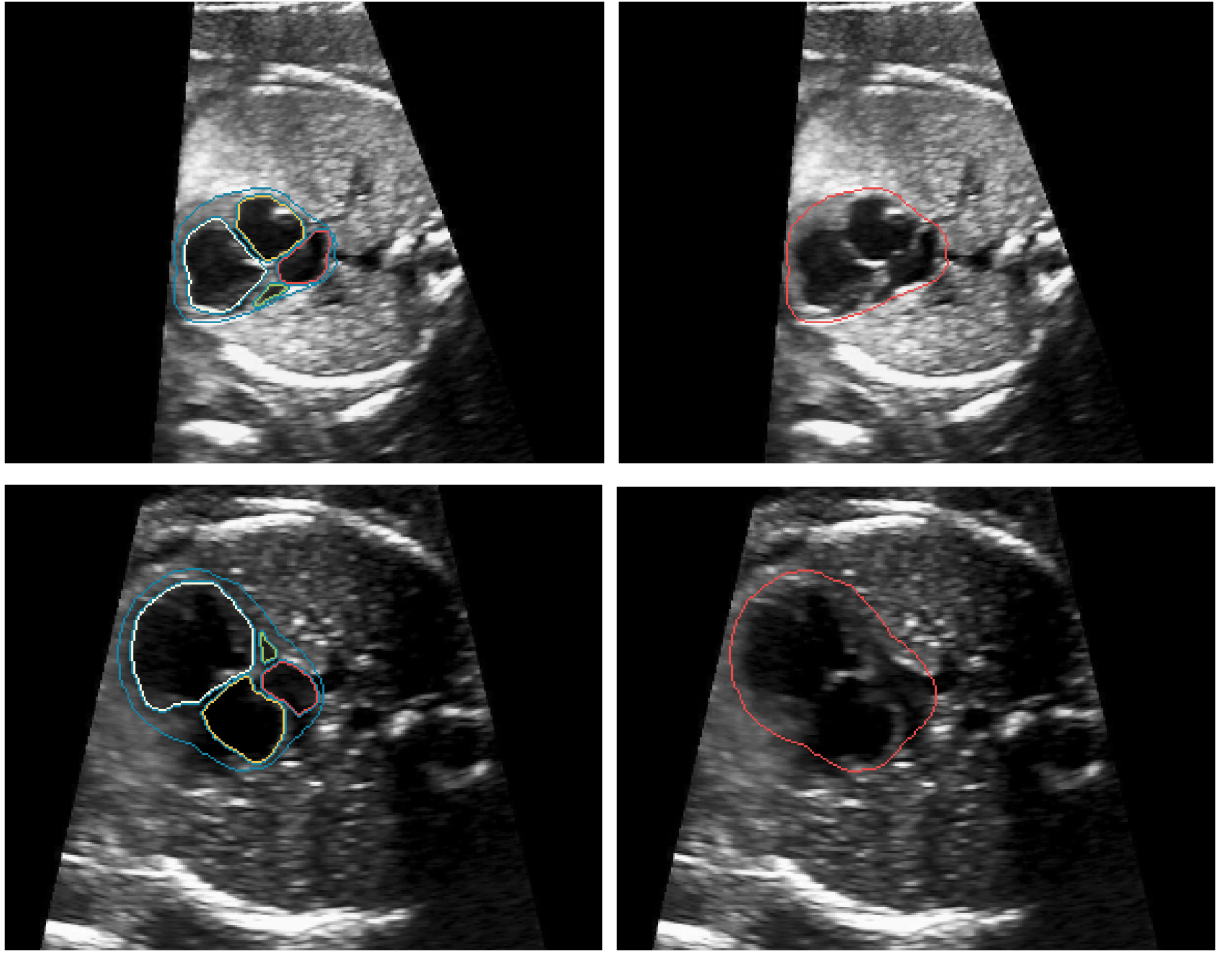

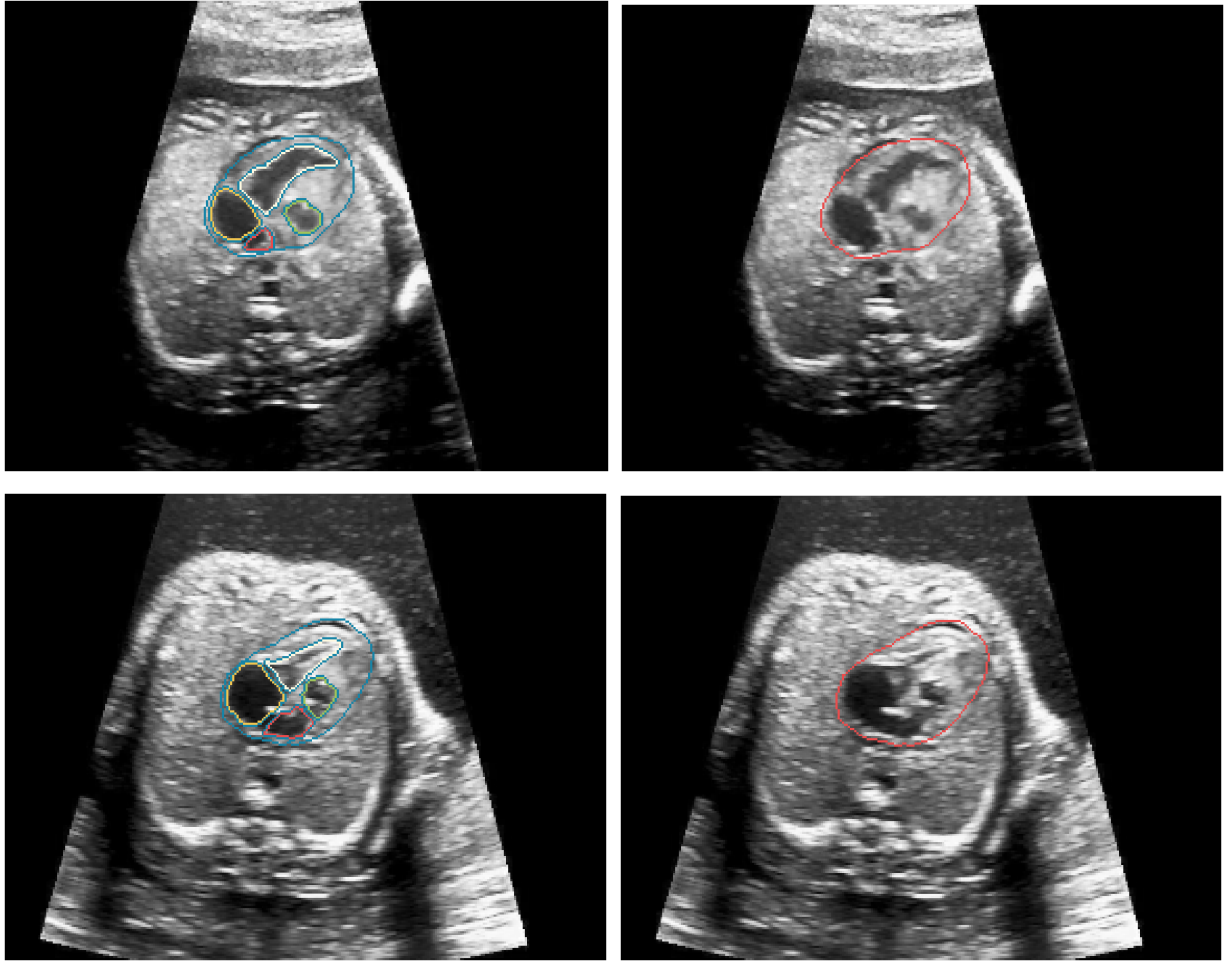

## Challenge 3: Split pane images

Several images in this dataset show two slightly different views of the heart side by side, in this case, only one of the hearts should be annotated. Usually one of the sides will show a clearer image of the heart, and this image should be annotated, i.e the heart in which it is easier to delineate between the chambers should be annotated (All five classes should be annotated).

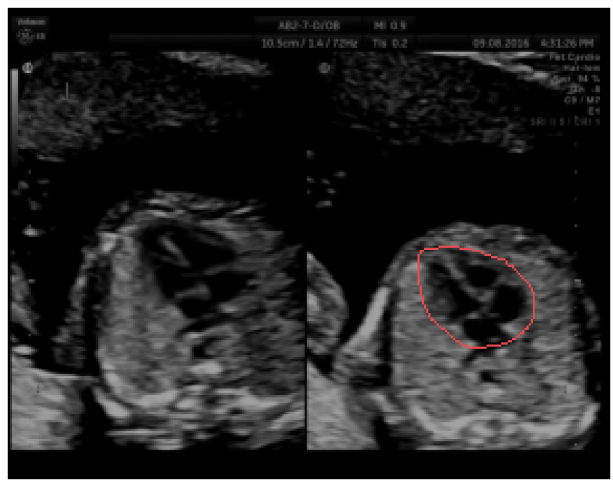

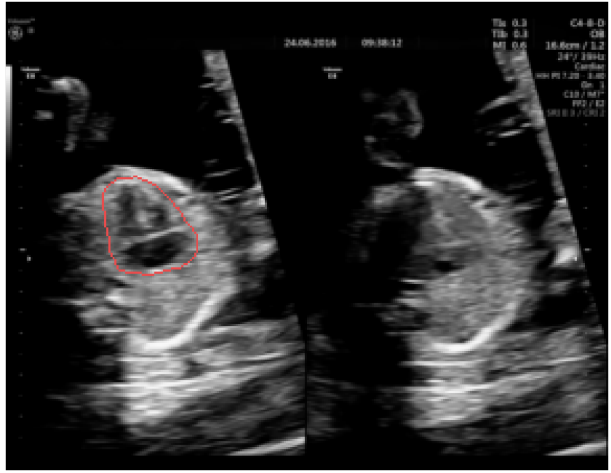

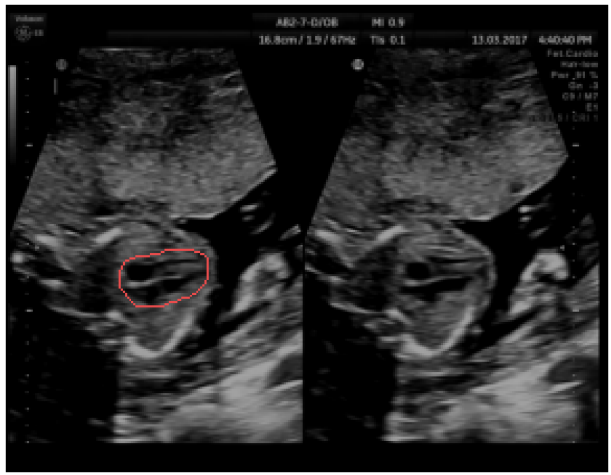

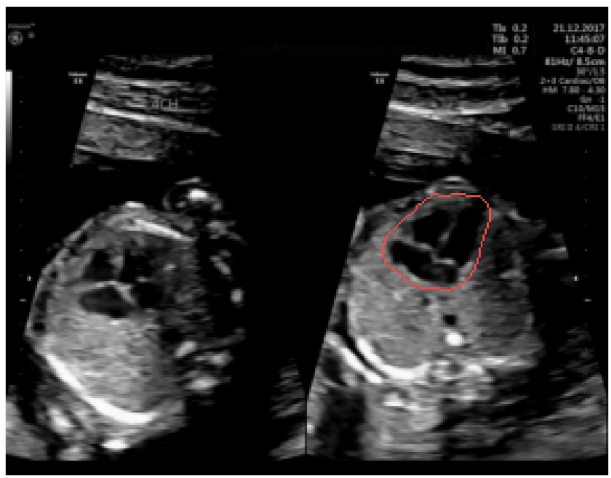

Thank you for your annotation efforts, and we want you to know that these annotations will be directly contributing to improving clinical care in fetal ultrasound screening.


\setlength{\voffset}{\originalVOffset}
\setlength{\hoffset}{\originalHOffset}
% \includepdf[fitpaper=true, pages=-]{Instructions.pdf}
\end{document}